\documentclass[misq]{informs3b} 

\OneAndAHalfSpacedXI 

\usepackage{natbib}
\bibpunct[, ]{(}{)}{,}{a}{}{,}%
\def\bibfont{\small}%

\usepackage[resetlabels]{multibib}
\newcites{sec}{References}

\TheoremsNumberedThrough     
\ECRepeatTheorems

\EquationsNumberedThrough    

\usepackage{caption}

\usepackage[bottom]{footmisc}
\usepackage{afterpage}

\usepackage{xcolor}

\usepackage[bookmarks, hidelinks]{hyperref}
\hypersetup{
	bookmarksnumbered=true,     
	bookmarksopen=true,
	bookmarksopenlevel=2,    
	colorlinks=true,
	linkcolor={red!50!black},
	citecolor={blue!50!black},
	urlcolor={blue!80!black},
}

\usepackage{bookmark}
\usepackage{etoolbox}
\usepackage{tocloft}
\makeatletter
\pretocmd\endfigure{%
	\addtocontents{lof}{\protect{%
			\bookmark[
			rellevel=1,
			keeplevel,
			dest=\@currentHref,
			]{Figure \thefigure: \@currentlabelname}}}%
	\bookmark[
	rellevel=1,
	keeplevel,
	dest=\@currentHref,
	]{Figure \thefigure: \@currentlabelname}%
}{}{\errmessage{Patching \noexpand\endfigure failed}}
\pretocmd\endtable{%
	\addtocontents{lof}{\protect{%
			\bookmark[
			rellevel=1,
			keeplevel,
			dest=\@currentHref,
			]{Table \thetable: \@currentlabelname}}}%
	\bookmark[
	rellevel=1,
	keeplevel,
	dest=\@currentHref,
	]{Table \thetable: \@currentlabelname}%
}{}{\errmessage{Patching \noexpand\endtable failed}}
\makeatother

\usepackage{tabularx}
\usepackage{cases}
\usepackage{threeparttable}

\usepackage{comment}

\usepackage{algorithm} 
\usepackage[noend]{algpseudocode} 
\usepackage{tikz}
\makeatletter
\newcount\dirtree@lvl 
\newcount\dirtree@plvl
\newcount\dirtree@clvl
\def\dirtree@growth{%
	\ifnum\tikznumberofcurrentchild=1\relax
	\global\advance\dirtree@plvl by 1
	\expandafter\xdef\csname dirtree@p@\the\dirtree@plvl\endcsname{\the\dirtree@lvl}
	\fi
	
	\global\advance\dirtree@lvl by 1\relax
	\dirtree@clvl=\dirtree@lvl
	\advance\dirtree@clvl by -\csname dirtree@p@\the\dirtree@plvl\endcsname
	\pgf@xa=1cm\relax
	\pgf@ya=-0.6cm\relax
	\pgf@ya=\dirtree@clvl\pgf@ya
	\pgftransformshift{\pgfqpoint{\the\pgf@xa}{\the\pgf@ya}}%
	
	\ifnum\tikznumberofcurrentchild=\tikznumberofchildren
	\global\advance\dirtree@plvl by -1
	\fi
}
\tikzset{
	dirtree/.style={
		growth function=\dirtree@growth,
		growth parent anchor=south west,
		parent anchor=south west,
		every child node/.style={anchor=west},
		edge from parent path={([xshift=2ex] \tikzparentnode\tikzparentanchor) 
			|- (\tikzchildnode\tikzchildanchor)},
	}
}
\makeatother

\newcommand\Tstrut{\rule{0pt}{2.6ex}}         

\usepackage[toc]{appendix}

\usepackage[nomessages]{fp}

\usepackage{enumitem}

\begin{document}

\newcolumntype{L}[1]{>{\raggedright\arraybackslash}p{#1}}
\newcolumntype{C}[1]{>{\centering\arraybackslash}p{#1}}
\newcolumntype{R}[1]{>{\raggedleft\arraybackslash}p{#1}}



\RUNTITLE{Assigning Industry Codes to Firms}

\begin{center}
\textbf{\Large Exploiting Expert Knowledge for Assigning Firms to Industries: A Novel Deep Learning Method}    
\end{center}

\hspace{0.2cm}
\begin{center}

Xiaohang Zhao$^1$, Xiao Fang$^{2,*}$, Jing He$^{2}$, Lihua Huang$^{3}$ \\
\vspace{0.2cm}
$^1$ School of Information Management \& Engineering, Shanghai University of Finance and Economics, Shanghai, China \\

$^2$ Lerner College of Business and Economics, University of Delaware, Newark, DE, USA \\

$^3$ School of Management, Fudan University, Shanghai, China \\

\vspace{0.1cm}
$*$ Corresponding Author: Xiao Fang, \href{mailto:xfang@udel.edu}{xfang@udel.edu} 
\end{center}
\vspace{0.3cm}

\noindent \textbf{Abstract:} Industry assignment, which assigns firms to industries according to a predefined Industry Classification System (ICS), is fundamental to a large number of critical business practices, ranging from operations and strategic decision making by firms to economic analyses by government agencies. Three types of expert knowledge are essential to effective industry assignment: definition-based knowledge (i.e., expert definitions of each industry), structure-based knowledge (i.e., structural relationships among industries as specified in an ICS), and assignment-based knowledge (i.e., prior firm-industry assignments performed by domain experts). Existing industry assignment methods utilize only assignment-based knowledge to learn a model that classifies unassigned firms to industries, and overlook definition-based and structure-based knowledge. Moreover, these methods only consider which industry a firm has been assigned to, but ignore the time-specificity of assignment-based knowledge, i.e., when the assignment occurs. To address the limitations of existing methods, we propose a novel deep learning-based method that not only seamlessly integrates the three types of knowledge for industry assignment but also takes the time-specificity of assignment-based knowledge into account. Methodologically, our method features two innovations: dynamic industry representation and hierarchical assignment. The former represents an industry as a sequence of time-specific vectors by integrating the three types of knowledge through our proposed temporal and spatial aggregation mechanisms. The latter takes industry and firm representations as inputs, computes the probability of assigning a firm to different industries, and assigns the firm to the industry with the highest probability. We conduct extensive evaluations with two widely used ICSs and demonstrate the superiority of our method over prevalent existing methods. 

\hspace{0.1cm} 


\hspace{1cm} 

\section{Introduction} \label{sec:intro}

Fostered by the rapid advancement of Information Technology (IT), Financial Technology (Fintech) has attracted increasing research attention from the business field in general and the Information Systems (IS) field in particular \citep{hendershott_call_2017, goldstein_fintech_2019}. Generally speaking, Fintech refers to the development of IT-based solutions to solve important financial problems with the goal of making financial services and business practices more efficient and effective \citep{hendershott_call_2017, goldstein_fintech_2019}. Over the years, IS researchers have tackled important financial problems ranging from discovering firms' financial risks to measuring firms' dyadic business proximity, using advanced IT solutions, especially solutions based on artificial intelligence and machine learning \citep{bao_simultaneously_2014, shi_toward_2016}. 
One important financial problem is to assign firms to industries according to a predefined Industry Classification System (ICS), namely the industry assignment problem \citep{wood_automated_2017}.
According to \cite{hoberg_text-based_2016}, ICSs that define industry boundaries and industry competitiveness are central to business and economics research. There are two streams of research dedicated to the design of ICSs \citep{wood_automated_2017}. One stream develops new ways of grouping firms and designs novel structures of ICSs that are different from widely used existing ones. 
The other stream, including our study, solves the industry assignment problem for existing ICSs.

An ICS is a taxonomy of business activities that aims to group firms operating similar lines of business into the same categories \citep{phillips_industry_2016}. The industry categories of an ICS are typically organized hierarchically as a tree, each node of which corresponds to an industry and is labeled by a numeric code as well as a title of the industry. The granularity of the classification increases from the top level of the tree to the bottom level. 
One prominent ICS is the North American Industry Classification System (NAICS), which was developed in collaboration by the Canadian, Mexican, and U.S. governments, and is widely used by government agencies, business practitioners, and academic researchers  \citep{bhojraj_whats_2003,phillips_industry_2016}. 
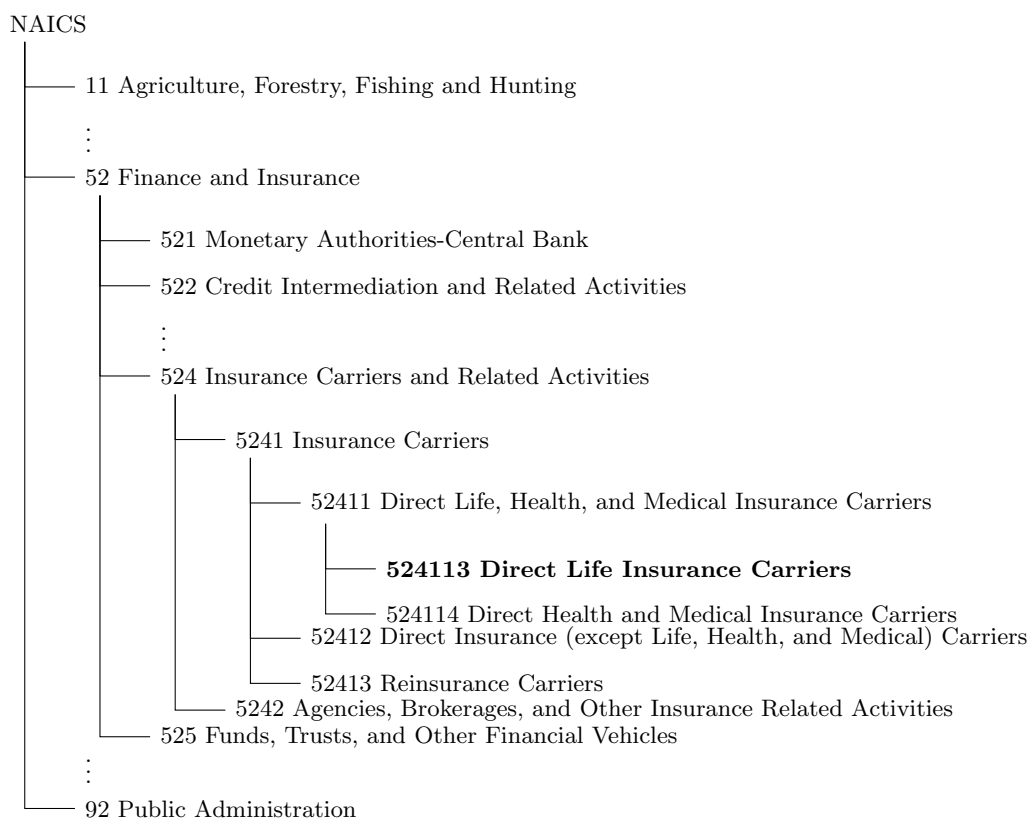
\begin{figure}[h]
	\centering
	\begin{tikzpicture}[dirtree]
		\tikzstyle{every node}=[font=\footnotesize]
		\node (root) {NAICS}
		child {node (11) {11 Agriculture, Forestry, Fishing and Hunting}}
		child {node {\vdots} edge from parent [draw=none]}
		child {
			node (52) {52 Finance and Insurance}
			child {node (521) {521 Monetary Authorities-Central Bank}}
			child  {node (522) {522 Credit Intermediation and Related Activities}}
			child {node {\vdots} edge from parent [draw=none]}
			child  {
				node (524) {524 Insurance Carriers and Related Activities}
				child {
					node (5241) {5241 Insurance Carriers}
					child {
						node (52411) {52411 Direct Life, Health, and Medical Insurance Carriers}
						child {node (524113) {\textbf{524113 Direct Life Insurance Carriers}}}
						child {node {524114 Direct Health and Medical Insurance Carriers}}}
					child {node {52412 Direct Insurance (except Life, Health, and Medical) Carriers}}
					child {node {52413 Reinsurance Carriers}}
				}
				child {node {5242 Agencies, Brokerages, and Other Insurance Related Activities}}
			}
			child {node (525) {525 Funds, Trusts, and Other Financial Vehicles}}
		}
		child {node {\vdots} edge from parent [draw=none]}
		child {
			node (92) {92 Public Administration}
		};
	\end{tikzpicture}
	\caption{A Partial View of the NAICS}
	\label{fig:example_ICS}
\end{figure}
Figure \ref{fig:example_ICS} provides an excerpt from the NAICS hierarchy that highlights the ``Direct Life Insurance Carriers'' industry, which is coded as 524113.\footnote{The tree is based on the NAICS taxonomy as revised in 2012. See \url{https://www.census.gov/naics/?58967?yearbck=2012} for the complete hierarchy.} As shown, the NAICS is comprised of five levels of industries. A root node is placed at level zero to denote the entire ICS. From the first level to the fifth level, the industry codes have two, three, four, five, and six digits, respectively.  
Given a target ICS, a focal level of interest, and the firms to be assigned, the objective of the industry assignment problem is to assign each firm to a focal-level industry of the target ICS that best covers the firm's business activities. As an example, when the target ICS is the NAICS  (Figure \ref{fig:example_ICS}) and the focal level is two, each firm needs to be assigned to one of the 99 NAICS second-level industries (e.g., 521 [Monetary Authorities-Central Bank]).

Industry assignment is necessary for an expanding array of important business practices.
For firms, knowing the industry to which a firm belongs is important for its daily operations and strategic decisions. For example, 
a firm needs to know its industry code to determine whether it is qualified for certain tax deduction programs or eligible for bidding on government contracts that are only offered to specific industries.\footnote{Please visit \url{https://www.census.gov/naics/} and then click on the ``FAQs'' tab.\label{fn:tax_deduction}} 
A firm also frequently needs to identify its industry peers, i.e., those firms that are assigned to the same industry as itself. These peers serve as references for the firm when making a series of strategic decisions, such as shaping corporate policies \citep{cao_peer_2019} and determining executive compensation \citep{bizjak_are_2011}. 
For investors, industry assignment information is often used for relative evaluation \citep[e.g.,][]{goodman_industry_1983}, where the stock value of a firm is evaluated within the context of its industry peers. 
For government agencies, industry assignment information has been used for collecting and analyzing economic data. As an example, the U.S. Bureau of Labor Statistics reports employment and output data for firms grouped by their two-digit NAICS codes,\footnote{\url{https://www.bls.gov/emp/tables/industry-employment-and-output.htm}\label{fn:employment_data}} 
which shed light on how technology development impacts labor productivity by industry.\footnote{\url{https://www.bls.gov/emp/frequently-asked-questions.htm}}  
Industry assignment also plays an important role in business research. 
Researchers often employ industry assignment information to identify economically comparable firms from the same industry as a control group, restrict samples to specific industries, develop industry dummies to control for fixed effects, or detect industry effects \citep{kahle_impact_1996, mcgahan_how_1997, cavaglia_increasing_2000,weiner_impact_2005}. Moreover, it is common to utilize industry assignment information for several of the purposes listed above in a single study, e.g., \cite{ahern_importance_2014} and \cite{bonaime_does_2018}.\footnote{As one example, \cite{ahern_importance_2014} investigate the propagation of M\&A activities through industry links. Their work is based on the input--output data from the Bureau of Economic Analysis, which are impossible to retrieve without knowing the assignment of firms' NAICS industries.}

Given the important role of industry assignment in business practices and research, substantial efforts have been devoted to it. 
Traditionally, industry assignment was partially a manual procedure conducted by human experts from authorities such as the U.S. Census Bureau, which is costly and time-consuming \citep{kearney_automated_2005, gweon_three_2017}.
Consequently, attempts have been made to develop methods for automatic industry assignment. Such methods typically learn a classification model from prior assignment cases performed by human experts and apply the learned model to automatically classify unassigned firms  \citep{kearney_automated_2005, rodrigues_automatic_2012, wood_automated_2017}. For example, \cite{kearney_automated_2005} classify firms into NAICS industries with keyword matching techniques, while \cite{wood_automated_2017} build a deep learning-based classification model that takes texts describing a firm as input and predicts the firm's NAICS industry.

Three types of expert knowledge are essential for effective industry assignment: definition-based, structure-based, and assignment-based knowledge. These types of knowledge are created by domain experts, who design an ICS or assign firms to industries according to an ICS, and they inform automatic industry assignment in different respects. Definition-based knowledge for an industry is a paragraph describing the business activities covered by the industry. 
As an example, the textual definition for the NAICS industry ``Direct Life Insurance Carriers'' (coded as 524113 in Figure \ref{fig:example_ICS}), is 
\begin{quote}
	\textit{This U.S. industry comprises establishments primarily engaged in initially underwriting (i.e., assuming the risk and assigning premiums) annuities and life insurance policies, disability income insurance policies, and accidental death and dismemberment insurance policies.} \footnote{ \url{https://www.census.gov/naics/?input=52&chart=2012&details=524113}}
	
\end{quote}
A deep grasp of definition-based knowledge allows an industry assignment method to form an informative expectation about which industry a firm should be assigned to, because this type of knowledge defines the scope of business activities covered by an industry. 
Structure-based knowledge refers to the hierarchical way of organizing industries. The hierarchy is presented as a tree structure (e.g., the NAICS structure illustrated in Figure \ref{fig:example_ICS}) and reveals how industries are related to each other in terms of the business activities they cover. \label{pg:nacis} 
An industry assignment method could leverage the knowledge of industry relatedness to infer which firms should be classified into an industry based on firms that have already been assigned to its related industries.
Assignment-based knowledge refers to prior firm--industry assignments by domain experts.
While definition-based knowledge pinpoints the core concepts shaping an industry, assignment-based knowledge instantiates these concepts and reflects the knowledge of domain experts who classify firms according to an ICS.

However, existing industry assignment methods rely solely on assignment-based knowledge to learn their industry assignment models while ignoring definition-based and structure-based knowledge \citep{kearney_automated_2005, rodrigues_automatic_2012, wood_automated_2017}. Furthermore, these methods adopt a static view of assignment-based knowledge, as if the entire assignment history were created at a single time point, while assignment-based knowledge is actually accumulated incrementally and evolves dynamically over time. 
Typically, the set of firms assigned to an industry changes over time because new firms might be added to the industry while existing firms might be reassigned to other industries or even removed from the firm universe. 
Therefore, in addition to definition-based knowledge,  the exact business activities covered by an industry also depend on the choices of human experts who keep fine-tuning their interpretation of the industry definition in response to continuous economic innovations.
Nevertheless, existing studies only consider which industry a firm has been assigned to, and ignore the time-specificity of assignment-based knowledge, i.e., when the assignment occurs.

It is challenging to design a method that simultaneously considers the three types of expert knowledge discussed above because of the heterogeneity of the data formats: structure-based knowledge is
encoded as a tree, definition-based knowledge is presented textually, and assignment-based knowledge is time-stamped. The central challenge is to develop an embedding space where all three types of expert knowledge can be properly represented and integrated for industry assignment.
To address the research gaps discussed above, we propose a novel deep learning-based method. 
In contrast to existing industry assignment methods, which rely solely on assignment-based knowledge, our method seamlessly integrates definition-based, structure-based, and assignment-based knowledge for industry assignment. Moreover, our method considers the time-specificity of assignment-based knowledge, which is neglected by existing industry assignment methods. In doing so, our study makes two methodological contributions: dynamic industry representation and hierarchical assignment. Dynamic industry representation embeds an industry as a sequence of time-specific vectors by integrating the three types of knowledge through our proposed temporal and spatial aggregation mechanisms.  Each vector represents the industry in a specific time period. Thus, dynamic industry representation distinguishes our method from existing industry assignment methods, which treat an industry as a static class label. Hierarchical assignment computes the time-specific probability of assigning a firm to a focal-level industry. It considers industries across the ICS hierarchy and incorporates structure-based knowledge into the probability computation, unlike existing industry assignment methods that only consider industries at the focal level and ignore structure-based knowledge. 

\section{Literature Review} \label{sec:rw}

Two streams of research are closely related to our study: existing methods for industry assignment and state-of-the-art machine and deep learning models that can be adapted to process structure-based or definition-based knowledge. In this section, we review each stream of related research and highlight the key novelties of our study.

\subsection{Fintech and Industry Assignment} \label{sec:rw:autoIA}
Our study generally falls into the research field of Fintech \citep{hendershott_call_2017, goldstein_fintech_2019}. In this field, IS researchers have actively developed advanced IT solutions to solve a diverse set of critical financial problems. For example, \cite{hu_network-based_2012} treat firms (banks) that are linked by their financial relationships as a network and develop a network-based method to analyze firms' financial risks. \cite{bao_simultaneously_2014} also study financial risks, but from a different perspective: they propose a topic model-based method to quantify firms' financial risks based on their textual risk disclosures. \cite{abbasi_metafraud:_2012} focus on financial fraud and propose a meta-learning framework to detect firms' financial frauds. 
We study the industry assignment problem and contribute to the Fintech field with a novel and effective solution method for this important problem.\footnote{The basic classification unit of an ICS is usually a firm. In some cases, a smaller classification unit might be used, e.g., a single factory among many factories operated by a firm. Our method is still applicable to classify these smaller units if their business descriptions are provided.}
A research stream related to the industry assignment problem attempts to design new ICSs by clustering firms with similar economical characteristics into homogeneous groups \citep{jaffe_technological_1986, lee_search-based_2015, hoberg_text-based_2016, yang_automatic_2016, gao_exploring_2020}.
For example, \cite{jaffe_technological_1986} represents each firm as a vector of its distribution over patent classes and computes pairwise similarities between firms based on their corresponding vectors; \cite{lee_search-based_2015} measure the similarity between two firms based on how frequently they are co-searched on the EDGAR website; and 
\cite{hoberg_text-based_2016} represent each firm as a bag-of-words vector derived from its annual 10-K report, and compute the cosine similarity between firm vectors.
Different from these studies, our study tackles the industry assignment problem, which assigns firms into industries of an existing ICS and is essentially a classification problem.

Early methods use keyword matching techniques to automate industry assignment \citep{chen_error_1993, kearney_automated_2005, lim_automatic_2005, jung_web-based_2008}. For example, \cite{kearney_automated_2005} document a two-step approach that assigns NAICS industries to newly birthed firms based on information collected from their application forms for employer identification numbers. In the first step, the words in an application form are matched against a set of keyword dictionaries, each of which corresponds to an industry and is constructed from historical firm--industry assignments. This step generates a set of candidate industries for a firm. The second step employs a logistic regression to select the most probable industry from the candidates. These methods rely on the quality of the keyword dictionaries, the construction of which is labor-intensive. Thus, they do not scale to ICSs with a large number of industries. 

Modern industry assignment methods utilize more sophisticated machine learning models, thereby avoiding the scalability issue \citep{roelands_classifying_2010, thompson_creating_2012, rodrigues_automatic_2012, gweon_three_2017, wood_automated_2017}. 
These methods formulate the industry assignment problem as a multi-class classification problem and learn a classification model from prior firm--industry assignments conducted by domain experts. The learned model can then be used to classify a new firm into one of the focal-level industries. For instance, \cite{roelands_classifying_2010} classify a firm into a top-level industry of the NACE (a European ICS) taxonomy by using texts from the firm's website as inputs. They experiment with several machine learning methods, including support vector machines (SVMs).  
Deep learning models have also been utilized for industry assignment \citep{wood_automated_2017, tagarev_comparison_2019, wei_abr-hic_2019}. \cite{wood_automated_2017} apply a multi-layer perceptron model to assign firms to NAICS industries. The model is trained with prior firm--industry assignments, and the input feature for a firm, or firm representation, is a high-dimensional, sparse bag-of-words vector encoding the textual information about the firm. 
Subsequent studies have investigated more sophisticated text representation methods for firm representation. 
Specifically, \cite{tagarev_comparison_2019} compare four methods of representing firms' description documents as input features for industry assignment, and conclude that the ULMFiT method \citep{howard_universal_2018} achieves the best industry assignment performance. 
\cite{wei_abr-hic_2019} propose a novel firm representation method by combining both textual and non-textual information about a firm via an attention-based recurrent neural network architecture. \cite{tagarev_comparison_2019} and \cite{wei_abr-hic_2019} focus on firm representation but ignore industry representation, and these two studies employ an existing classification method for firm-industry assignment.
Our study, on the other hand, develops novel approaches for industry representation and firm-industry assignment, but uses an existing method, i.e., Doc2Vec \citep{le_distributed_2014}, for firm representation.
As discussed in Section \ref{sec:intro}, definition-based, structure-based, and assignment-based knowledge are all essential for effective industry assignment. However, existing industry assignment methods neglect definition-based and structure-based knowledge, which dampens their classification performance. Recent developments in machine and deep learning provide models that can be adapted to process definition-based or structure-based knowledge for industry assignment; we review these models next.

\subsection{Hierarchical Classification and Label Embedding}  \label{sec:rw:class}

Hierarchical classification methods can be adapted to process structure-based knowledge.  
These methods use a tree-shaped label hierarchy (e.g., structure-based knowledge in our study) to organize a set of classifiers, each of which is contained in a non-leaf node of the tree and is only responsible for classifying an instance into one of the child classes of the node  \citep{koller_hierarchically_1997, mccallum_improving_1998, dekel_large_2004, weinberger_large_2009, silla_survey_2011}.\footnote{In the machine learning literature, classes in a classification model are more generally referred to as labels.} The classifier at a non-leaf node is learned from training instances whose labels belong to the branches originating from the node. 
When classifying an instance, the root classifier first predicts it as one of the first-level classes. Next, the classifier at the node of the predicted class is invoked to predict the instance's class at the next level. This step is repeated until a leaf-level class is predicted. Alternatively, a tree-shaped label hierarchy can be used to compose vector representations of classes \citep{dekel_large_2004, weinberger_large_2009}. For example, \cite{dekel_large_2004} propose representing a class as the summation of vectors representing its ascendant classes in the label hierarchy. These representation vectors are learned from training instances and are then employed to predict the class of an unassigned instance.

Label embedding methods are suitable for leveraging definition-based knowledge. 
In a label embedding method, each label or instance is described by a textual document (i.e., a sequence of words) and represented by a numeric vector that captures the semantics of the document \citep{yazdani_model_2015, nam_all-text_2016, pappas_gile:_2019}. \cite{yazdani_model_2015} propose representing a label document by concatenating its component word vectors and encoding an instance document with the summation of its component word vectors, where the word vectors are learned using a pretrained word embedding model. \cite{nam_all-text_2016} apply a document embedding model proposed by \cite{le_distributed_2014} to encode both label and instance documents. In \cite{pappas_gile:_2019}, a label document is represented as the average of its component word vectors, whereas an instance document is encoded using a sequential compositional neural network.
The compatibility score between an instance--label pair is computed by applying a matching function to their representation vectors. The matching function can be a simple inner product of label and instance vectors \citep{yazdani_model_2015} or a more complicated neural network that takes label and instance vectors as inputs \citep{pappas_gile:_2019}. Finally, the parameters of a label embedding model are learned in such a way that the correct label for a training instance has the highest compatibility score among all of the labels. By treating industry definitions as label documents, we can adapt label embedding methods to process definition-based knowledge.



\subsection{Key Novelties of Our Method}\label{sec:rw:novelties}

Our literature review suggests several research gaps. Existing industry assignment methods only consider assignment-based knowledge but ignore the time specificity of this knowledge; they also neglect definition-based and structure-based knowledge. While hierarchical classification methods can be adapted to process structure-based knowledge, these methods are not designed to leverage definition-based knowledge. On the other hand, label embedding methods are suitable for processing definition-based knowledge, but they do not work with structure-based knowledge. Moreover, both hierarchical classification and label embedding methods neglect the time specificity of assignment-based knowledge. Our method simultaneously considers the three types of knowledge as well as the time specificity of assignment-based knowledge. Therefore, our method is distinct from existing methods, as summarized in Table \ref{tb:methodcomp}. Methodologically, our method features two innovations: 
dynamic industry representation and hierarchical assignment. Dynamic industry representation embeds an industry as a sequence of time-specific vectors, in contrast to existing industry assignment methods and hierarchical classification methods that represent an industry as a static class label. It derives these time-specific vectors by leveraging the three types of knowledge. Label embedding methods, in contrast, represent an industry as a static vector based solely on definition-based knowledge. Furthermore, hierarchical assignment distinguishes our method from existing methods in that it incorporates structure-based knowledge into the computation of the time-specific probability of assigning a firm to an industry.   

\begin{table}[h]
	\caption{Comparison between Our Method and Existing Methods in Terms of the Used Knowledge}
	\label{tb:compare_methods}
	\centering
	\begin{threeparttable}
		\begin{tabular}{L{0.35\textwidth} C{0.17\textwidth} C{0.17\textwidth} C{0.18\textwidth}}
			\hline
			& Definition-based Knowledge 
			& Structure-based Knowledge 
			& Assignment-based Knowledge\tnote{*} \\ \hline
			Existing Industry Assignment Methods, e.g., \cite{wood_automated_2017, tagarev_comparison_2019} & No & No & Static \Tstrut \\ 
			Hierarchical Classification Methods, e.g., \cite{silla_survey_2011} & No & Yes & Static \Tstrut \\ 
			Label Embedding Methods, e.g., \cite{pappas_gile:_2019}	& Yes & No & Static \Tstrut \\  
			Our Method	& Yes  & Yes & Dynamic \Tstrut \\ \hline
		\end{tabular} 
		\begin{tablenotes}\footnotesize
			\item[*] For the industry assignment problem, an instance of firm-industry assignment consists of three pieces of information: a firm, the industry the firm was assigned to, and the timestamp when the assignment occurred. An industry assignment method is based on static assignment-based knowledge if it only considers the assigned firm-industry pair but ignores the time-specificity of the assignment, while it is based on dynamic assignment-based knowledge if it utilizes all three pieces of information.
		\end{tablenotes}
	\end{threeparttable}
	\label{tb:methodcomp}
\end{table}

\section{Problem Formulation} \label{sec:probformulation}

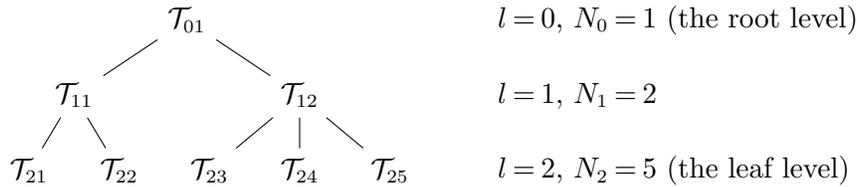
\begin{figure}[h]
	\centering 
	\begin{tikzpicture}
		\tikzstyle{level 1}=[sibling distance=3cm, level distance=1cm]
		\tikzstyle{level 2}=[sibling distance=1.2cm, level distance=1cm]
		\node (root) {${\mathcal{T}}_{01}$}
		child {
			node (11) {${\mathcal{T}}_{11}$}
			child {
				node (21) {${\mathcal{T}}_{21}$}
			}
			child {
				node (22) {${\mathcal{T}}_{22}$}
			}
		}
		child {
			node (12) {${\mathcal{T}}_{12}$}
			child {
				node (23) {${\mathcal{T}}_{23}$}
			}
			child {
				node (24) {${\mathcal{T}}_{24}$}
			}
			child {
				node (25) {${\mathcal{T}}_{25}$}
			}
		};
		\node[anchor=west] (l0) at (4, 0) {$l=0$, $N_0=1$ (the root level)};
		\node[anchor=west] (l1) at (4, -1) {$l=1$, $N_1=2$};
		\node[anchor=west] (l2) at (4, -2) {$l=2$, $N_2=5$ (the leaf level)};
	\end{tikzpicture}
	\caption{Industry Tree for an Example ICS ($N=7$, $L=2$, \textmd{excluding} ${\mathcal{T}}_{01}$)}
	\label{fig:ic_tree}
\end{figure}

We consider a target ICS in which the industries are organized hierarchically as an ``industry tree''. Figure \ref{fig:ic_tree} shows the industry tree for an example ICS that will be used throughout this paper.
In general, the $i$th industry at level $l$ is denoted as ${\mathcal{T}}_{li}$, assuming an arbitrary ordering of the industries at the level. 
Let $N_l$ be the number of industries at level $l$. At the root level, $l=0$ and $N_0=1$, which means that ${\mathcal{T}}_{01}$ is the single root node that represents the entire ICS. At the leaf level, $l=L$, which indicates that the industry tree contains $L$ levels in total without counting the root level. The set of industries at level $l$ is written as
$${\mathcal{T}}_l=\{ {\mathcal{T}}_{l1}, {\mathcal{T}}_{l2}, \dots, {\mathcal{T}}_{lN_l} \},$$ 
and the complete set of industries is ${\mathcal{T}}= \cup_{l=1}^{L} {\mathcal{T}}_{l}$. The total number of industries is computed as $N=\sum_{l=1}^{L} N_l$, excluding the root node. The target ICS is applied to a firm universe ${\mathcal{U}}$. 

Given a focal level of interest $l^{*}$, $1 \le l^{*} \le L$, there exists a set of firm--industry assignment cases that are accumulated in periods $1:T$, which is short for $[1, 2, \dots, T]$. Each assignment case is a tuple $(j, y^{(t, j)})$ comprised of a firm and its industry assigned in a particular period, where $j\in {\mathcal{U}}$ represents a firm and $ y^{(t, j)}$ is the industry at level $l^{*}$ assigned to firm $j$ in period $t\in 1:T$. Let $doc^{(t,j)}$ denote the document describing the business of firm $j$ in period $t$.
We can now formally define the industry assignment problem: 

\begin{definition}[\textbf{Industry Assignment Problem}]
	\label{def:prob}
	Given a target ICS in which the industries ${\mathcal{T}}$ are organized as a tree, a corpus of textual definition for each industry in ${\mathcal{T}}$, a firm universe ${\mathcal{U}}$, its associated corpus $\{doc^{(t,j)} \ | \ j \in {\mathcal{U}}, \ t \in 1:T \}$
	of firms' business description documents from period $1$ to $T$,
	a focal industry level $l^{*}$, and a set of past firm--industry assignment cases performed by domain experts
	\begin{equation}
		\label{eq:dataset}
		{\mathcal{D}}=\{ (j, y^{(t, j)}) \ | \ j \in {\mathcal{U}}, \ y^{(t, j)} \in {\mathcal{T}}_{l^{*}}, \ t \in 1:T \},
	\end{equation}
	build a model to classify each unassigned firm $j' \in {\mathcal{U}}$ in period $T+1$ to an industry in ${\mathcal{T}}_{l^{*}}$ based on the business description document $doc^{(T+1,j')}$ of the firm in period $T+1$, such that the assigned industry best covers the business activities of the firm, which can be evaluated by comparing the industry $\hat{y}^{(T+1, j')}$ assigned by the model and the ground truth industry $y^{(T+1, j')}$ assigned by domain experts over all unassigned firms according to some metrics of interest.
\end{definition}

\section{DeepIA: A Deep Learning Method for Industry Assignment}\label{sec:method}
We present the three building blocks of DeepIA in Sections \ref{sec:method:encode_know}, \ref{sec:method:dir}, and \ref{sec:method:ha} and then introduce DeepIA in Section \ref{sec:method:learn}. The first building block encodes the three types of expert knowledge and represents firms; the second represents industries based on the encoded knowledge; and the third takes firm and industry representations as inputs and computes the probability of assigning a firm to an industry. The last two building blocks constitute main methodological novelties of our study and we highlight these novelties at the end of Sections \ref{sec:method:dir} and \ref{sec:method:ha}. For the convenience of the reader, we summarize important notation in Table \ref{tb:notation}.

\begin{table}[h!]
	\caption{Notation}
	\label{tb:notation}
	\begin{center}
		\begin{tabular}{L{0.1\textwidth} L{0.6\textwidth}}
			\hline 
			Notation &  \multicolumn{1}{c}{Description} \\ 
			\hline 
			${\mathcal{T}}_{li}$ & The $i$th industry at level $l$ \Tstrut \\ 
			${\mathcal{T}}_{li}^{(D)}$ & Definition-based knowledge of ${\mathcal{T}}_{li}$, Equation \ref{eq:definition_knowledge} \Tstrut \\ 
			${\mathcal{T}}_{li}^{(S)}$ & Structure-based knowledge of ${\mathcal{T}}_{li}$, Equation \ref{eq:structure_knowledge} \Tstrut \\ 
			${\mathcal{T}}_{li}^{(A,t)}$ & Assignment-based knowledge of ${\mathcal{T}}_{li}$ in period $t$, Equation \ref{eq:assign_knowledge} \Tstrut \\ 
			${\mathcal{P}}$ & Ancestor seeking operator, Definition \ref{def:operators} \Tstrut \\
			${\mathcal{C}}$ & Descendant seeking operator, Definition \ref{def:operators} \Tstrut \\		
			$x^{(t,j)}$ & Representation of firm $j$ in period $t$, Equation \ref{eq:firm_repr} \Tstrut \\
			$v_{li}^{(t)}$ & Dynamic representation of industry ${\mathcal{T}}_{li}$ in period $t$, Equation \ref{eq:temp_agg_simp} \Tstrut \\
			$P({\mathcal{T}}_{li}| j,t)$ &  Probability that firm $j$ is assigned to industry ${\mathcal{T}}_{li}$ in period $t$ among all the industries at level $l$, Equation \ref{eq:p_abs_upwards}  \Tstrut \\
			\hline  
		\end{tabular} 
	\end{center}
\end{table}

\subsection{Encoding Expert Knowledge and Representing Firms} \label{sec:method:encode_know}

We use ${\mathcal{T}}_{li}^{(D)}$, ${\mathcal{T}}_{li}^{(S)}$, and ${\mathcal{T}}_{li}^{(A, t)}$, respectively, to denote the definition-based, structure-based, and assignment-based knowledge of industry ${\mathcal{T}}_{li}$. It is straightforward to store the paragraph defining an industry as a sequence of words
\begin{equation}
	{\mathcal{T}}_{li}^{(D)} = <w_1^{(li)}, w_2^{(li)}, \dots, w_k^{(li)}, \dots>,
	\label{eq:definition_knowledge}
\end{equation}
where $w_k^{(li)}$ is the $k$th word in the paragraph defining industry ${\mathcal{T}}_{li}$. To express  structure-based knowledge, we define two operators that apply to the industry tree of an ICS:

\begin{definition}[\textbf{Ancestor- and Descendant-Seeking Operators}] Let ${\mathcal{P}}$ be the ancestor-seeking operator such that ${\mathcal{P}}^{k}({\mathcal{T}}_{li})$ returns the unique ancestor industry of ${\mathcal{T}}_{li}$ at level $l-k$ (i.e., $k$ levels above level $l$) for $1 \le l \le L$ and $1 \le k \le l$, and ${\mathcal{P}}^{0}({\mathcal{T}}_{li})$ returns ${\mathcal{T}}_{li}$ itself.
	Similarly, let ${\mathcal{C}}$ be the descendant-seeking operator such that ${\mathcal{C}}^{k}({\mathcal{T}}_{li})$ returns the set of descendant industries of ${\mathcal{T}}_{li}$ at level $l+k$ (i.e., $k$ levels below level $l$) for $0 \le l \le L-1$ and $1 \le k \le L-l$, and ${\mathcal{C}}^{0}({\mathcal{T}}_{li})$ returns ${\mathcal{T}}_{li}$ itself.
	\label{def:operators} 
\end{definition}

For simplicity, ${\mathcal{P}}({\mathcal{T}}_{li})$ means ${\mathcal{P}}^{1}({\mathcal{T}}_{li})$, or the parent industry of ${\mathcal{T}}_{li}$, and ${\mathcal{C}}({\mathcal{T}}_{li})$ means ${\mathcal{C}}^{1}({\mathcal{T}}_{li})$, or the set of child industries of ${\mathcal{T}}_{li}$. Using the defined operators, the structure-based knowledge of industry ${\mathcal{T}}_{li}$ is comprised of all of its ancestor and descendant industries (excluding the root node):
\begin{equation}
	{\mathcal{T}}_{li}^{(S)} = \begin{cases}
		\{ {\mathcal{C}}({\mathcal{T}}_{li}), {\mathcal{C}}^{2}({\mathcal{T}}_{li}), \dots, {\mathcal{C}}^{L-l}({\mathcal{T}}_{li}) \} \ &\text{if }  l = 1 \\
		\{ {\mathcal{P}}({\mathcal{T}}_{li}), {\mathcal{P}}^{2}({\mathcal{T}}_{li}), \dots, {\mathcal{P}}^{l-1}({\mathcal{T}}_{li}), {\mathcal{C}}({\mathcal{T}}_{li}), {\mathcal{C}}^{2}({\mathcal{T}}_{li}), \dots, {\mathcal{C}}^{L-l}({\mathcal{T}}_{li}) \} \ &\text{if } 2 \le l \le L-1 \\
		\{ {\mathcal{P}}({\mathcal{T}}_{li}), {\mathcal{P}}^{2}({\mathcal{T}}_{li}), \dots, {\mathcal{P}}^{l-1}({\mathcal{T}}_{li}) \} \ &\text{if } l = L \\
	\end{cases}
	\label{eq:structure_knowledge}
\end{equation}

\begin{example}
	Consider the target ICS in Figure \ref{fig:ic_tree}. We have  ${\mathcal{P}}({\mathcal{T}}_{25})=\{{\mathcal{T}}_{12}\}$ 
	and ${\mathcal{C}}({\mathcal{T}}_{12})=\{ {\mathcal{T}}_{23}, {\mathcal{T}}_{24}, {\mathcal{T}}_{25}\}$.
	By Equation \ref{eq:structure_knowledge}, ${\mathcal{T}}_{12}^{(S)}=\{{\mathcal{C}}({\mathcal{T}}_{12})\}=\{ {\mathcal{T}}_{23}, {\mathcal{T}}_{24}, {\mathcal{T}}_{25} \}$ and ${\mathcal{T}}_{25}^{(S)}=\{{\mathcal{P}}({\mathcal{T}}_{25})\}=\{ {\mathcal{T}}_{12}\}$.
\end{example}

The assignment-based knowledge of industry ${\mathcal{T}}_{li}$ in period $t$, denoted as ${\mathcal{T}}_{li}^{(A, t)}$, is the set of firms assigned to the industry in that period. Given past firm--industry assignments ${\mathcal{D}}$, we know which firm is assigned to which industry at focal level $l^{*}$ in which time period. Note that a firm is assigned to an industry if it is assigned to one of the industry's descendants at level $l^{*}$. Therefore, ${\mathcal{T}}_{li}^{(A, t)}$ consists of firms whose assigned industry in period $t$ belongs to ${\mathcal{T}}_{li}$'s descendants at level $l^{*}$. Formally, it is given by
\begin{equation}
	\label{eq:assign_knowledge}
	{\mathcal{T}}_{li}^{(A, t)} = \{ j \ | \ (j, y^{(t, j)}) \in {\mathcal{D}} ,  y^{(t, j)} \in {\mathcal{C}}^{l^{*}-l} ({\mathcal{T}}_{li} ) \}.
\end{equation}
By Definition \ref{def:operators}, ${\mathcal{C}}^{l^{*}-l} ({\mathcal{T}}_{li} )$ returns the set of ${\mathcal{T}}_{li}$'s descendant industries at focal level $l^*$ if $l<l^{*}$ and ${\mathcal{C}}^{l^{*}-l} ({\mathcal{T}}_{li} )={\mathcal{C}}^{0} ({\mathcal{T}}_{li} )=\{{\mathcal{T}}_{li}\}$ if $l=l^{*}$. Note that if $l>l^{*}$, firms assigned to industry ${\mathcal{T}}_{li}$ cannot be inferred from ${\mathcal{D}}$. Hence, we can only obtain assignment-based knowledge for industries at level $l$ for $l \leq l^{*}$.

\begin{example}
	Consider the target ICS in Figure \ref{fig:ic_tree} and focal level $l^{*}=2$. Past firm--industry assignments ${\mathcal{D}}$ in time periods $t=1,2$ are as follows:
	\begin{equation*}
		\begin{aligned}
			{\mathcal{D}}=&\{ (B, y^{(1, B)}={\mathcal{T}}_{21}), (C, y^{(1, C)}={\mathcal{T}}_{23}), (D, y^{(1, D)}={\mathcal{T}}_{25}), (E, y^{(1, E)}={\mathcal{T}}_{22}), \\
			& \ \ (B, y^{(2, B)}={\mathcal{T}}_{21}), (C, y^{(2, C)}={\mathcal{T}}_{25}), (D, y^{(2, D)}={\mathcal{T}}_{25}), (F, y^{(2, F)}={\mathcal{T}}_{24})\}.
		\end{aligned}
	\end{equation*}
	In ${\mathcal{D}}$, each uppercase letter represents a firm, and a tuple such as $(B, y^{(1, B)}={\mathcal{T}}_{21})$ indicates that firm $B$ is assigned to industry ${\mathcal{T}}_{21}$ in period 1. According to ${\mathcal{D}}$, firms $B, C, D, E$ are assigned to industries ${\mathcal{T}}_{21}, {\mathcal{T}}_{23}, {\mathcal{T}}_{25}, {\mathcal{T}}_{22}$, respectively, in period 1. In period 2, firms $B$ and $D$ stay in the same industry as period 1, while firm $C$ switches from ${\mathcal{T}}_{23}$ to ${\mathcal{T}}_{25}$. In addition, firm $E$ is removed from the firm universe in period $2$, while firm $F$ is newly added to the universe in period $2$ and is assigned to industry ${\mathcal{T}}_{24}$. 
	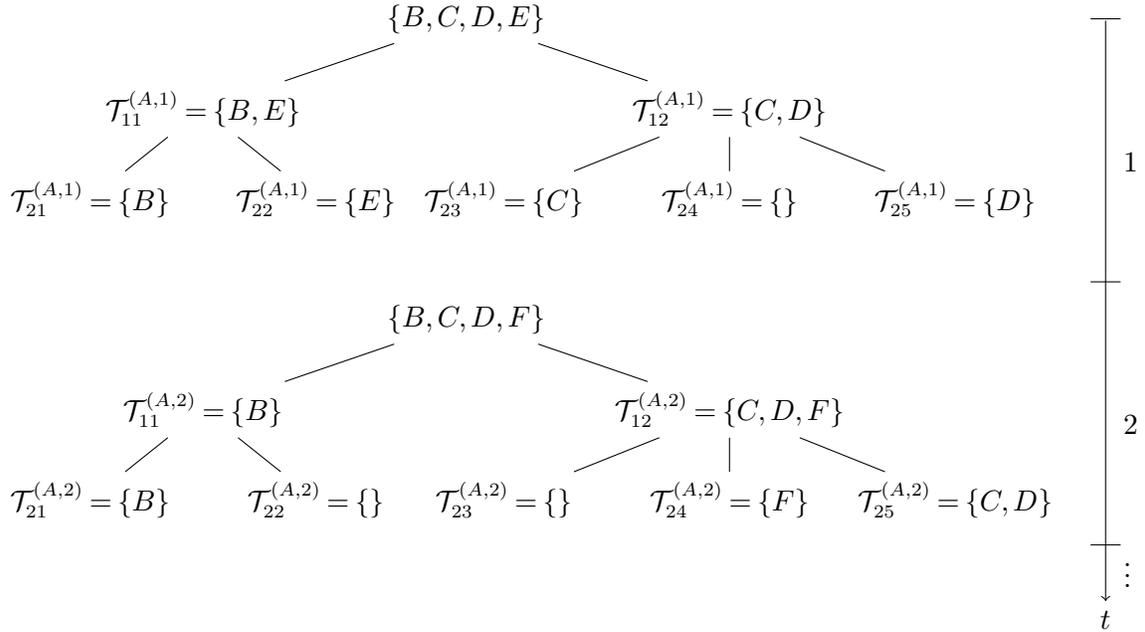
\begin{figure}[h]
		\centering 
		\begin{tikzpicture}
			\tikzstyle{level 1}=[sibling distance=7cm, level distance=1.2cm]
			\tikzstyle{level 2}=[sibling distance=3cm, level distance=1.2cm]
			\node (roott0) at (0,0) {$\{B,C,D,E\}$} 
			child {
				node (11t0) {${\mathcal{T}}_{11}^{(A, 1)}=\{B, E\}$}
				child {
					node (21t0) {${\mathcal{T}}_{21}^{(A, 1)}=\{B\}$}
				}
				child {
					node (22t0) {${\mathcal{T}}_{22}^{(A, 1)}=\{E\}$}
				}
			}
			child {
				node (12t0) {${\mathcal{T}}_{12}^{(A, 1)}=\{C,D\}$}
				child {
					node (23t0) {${\mathcal{T}}_{23}^{(A, 1)}=\{C\}$}
				}
				child {
					node (24t0) {${\mathcal{T}}_{24}^{(A, 1)}=\{\}$}
				}
				child {
					node (25t0) {${\mathcal{T}}_{25}^{(A, 1)}=\{D\}$}
				}
			};
			\node (roott1) at (0,-4) {$\{B,C,D,F\}$}
			child {
				node (11t1) {${\mathcal{T}}_{11}^{(A, 2)}=\{B\}$}
				child {
					node (21t1) {${\mathcal{T}}_{21}^{(A, 2)}=\{B\}$}
				}
				child {
					node (22t1) {${\mathcal{T}}_{22}^{(A, 2)}=\{\}$}
				}
			}
			child {
				node (12t1) {${\mathcal{T}}_{12}^{(A, 2)}=\{C,D,F\}$}
				child {
					node (23t1) {${\mathcal{T}}_{23}^{(A, 2)}=\{\}$}
				}
				child {
					node (24t1) {${\mathcal{T}}_{24}^{(A, 2)}=\{F\}$}
				}
				child {
					node (25t1) {${\mathcal{T}}_{25}^{(A, 2)}=\{C,D\}$}
				}
			};
			\node[inner sep=0pt] (t0) at (8.5, 0) {};
			\node (t1) at (8.5, -3.5) {};
			\node (tn) at (8.5, -8) {$t$};
			\draw[->] (t0) edge (tn);
			\draw (8.3, 0) -- (8.7, 0);
			\node[anchor=west] at (8.6, -1.9) {$1$};
			\draw (8.3, -3.5) -- (8.7, -3.5);
			\node[anchor=west] at (8.6, -5.4) {$2$};
			\draw (8.3, -7) -- (8.7, -7);
			\node[anchor=west] at (8.6, -7.3) {$\vdots$};
		\end{tikzpicture}
		\caption{Assignment-based Knowledge over Two Time Periods}
		\label{fig:ic_tree_assign}
	\end{figure}
	Figure \ref{fig:ic_tree_assign} illustrates the assignment-based knowledge derived from ${\mathcal{D}}$. As shown, the assignment-based knowledge ${\mathcal{T}}_{21}^{(A, 1)}$ of industry ${\mathcal{T}}_{21}$ in period 1 consists of firm $B$ assigned to the industry in that period, i.e., ${\mathcal{T}}_{21}^{(A, 1)}=\{B\}$. Similarly, we have ${\mathcal{T}}_{22}^{(A, 1)}=\{E\}$. By Equation \ref{eq:assign_knowledge}, the assignment-based knowledge ${\mathcal{T}}_{11}^{(A, 1)}$ of industry ${\mathcal{T}}_{11}$ in period 1 contains firms assigned to its descendant industries (i.e., ${\mathcal{T}}_{21}$ and ${\mathcal{T}}_{22}$) in that period, and hence ${\mathcal{T}}_{11}^{(A, 1)}=\{B, E\}$. Assignment-based knowledge is time-specific, thereby capturing the dynamics of industry assignment. For instance, the assignment-based knowledge of industry ${\mathcal{T}}_{11}$ changes from $\{B, E\}$ in period 1 to $\{B\}$ in period 2, as firm $E$ is removed from the firm universe in period 2. 
\end{example}

We embed firm $j$ in period $t$ as a $d$-dimensional vector $x^{(t, j)} \in R^{d}$. The representation of firm $j$ is time-specific because it should reflect the most up-to-date business activities of the firm.
We derive $x^{(t, j)}$ from $doc^{(t,j)}$, the business description document for firm $j$ in period $t$.\footnote{A firm's business description document might come from different sources. It can be extracted from the firm's website \citep{wood_automated_2017} or the firm's annual financial report (i.e., a 10-K report) \citep{pierre_automated_2001}. In general, such a document describes the business activities in which the firm engages and is therefore an ideal textual source for industry assignment.} 
To this end, we employ a document embedding model (DEM), e.g., the Doc2Vec model developed by \cite{le_distributed_2014}, which summarizes the semantics of each input document as a numeric vector. Specifically, we have 
\begin{equation}
	\label{eq:firm_repr}
	x^{(t,j)} = e_f\big( \text{DEM}(doc^{(t,j)}) \big),
\end{equation}
where DEM denotes a document embedding model and $e_f$ is a transformation layer. 
Because business description documents are typically long, it is computationally challenging to learn the parameters for a DEM as part of the model.
Therefore, following \cite{pappas_gile:_2019}, the DEM in Equation \ref{eq:firm_repr} is pretrained.
Because the DEM is trained without using the firm-industry assignment information, the resulted document vector might contain information that is irrelevant for industry assignment. To filter out the irrelevant information, we employ a transformation layer $e_f$ (e.g., a multi-layer perceptron) to extract informative firm representation $x^{(t,j)}$ from $\text{DEM}(doc^{(t,j)})$, where ``informative'' means ``informative for industry assignment''.
The parameters for $e_f$ are learned as part of the model parameters. The choices of the DEM and $e_f$ depend on the characteristics of the input documents (e.g., document length), and hence we specify them when reporting our experiments in Section \ref{sec:eval}. 

\subsection{Dynamic Industry Representation} \label{sec:method:dir}

The central idea of dynamic industry representation (DIR) is to represent an industry ${\mathcal{T}}_{li}$ as a sequence of time-specific vectors $\langle v_{li}^{(t)}\rangle$, $t=1,2,\dots,T$, 
each of which coherently integrates ${\mathcal{T}}_{li}$'s definition-based knowledge ${\mathcal{T}}_{li}^{(D)}$, structure-based knowledge ${\mathcal{T}}_{li}^{(S)}$, and assignment-based knowledge up to time period $t$: ${\mathcal{T}}_{li}^{(A, 1)}, {\mathcal{T}}_{li}^{(A, 2)}, \dots, {\mathcal{T}}_{li}^{(A, t)}$. 
To achieve this goal, ${\mathcal{T}}_{li}^{(D)}$ and ${\mathcal{T}}_{li}^{(A, t)}$ are respectively embedded as vectors $v_{li}^{(D)}\in R^{d}$ and $v_{li}^{(A, t)}\in R^{d}$, $t=1,2,\dots,T$. Next, embedding vectors $v_{li}^{(D)}$, $v_{li}^{(A, 1)}, v_{li}^{(A, 2)}, \dots ,v_{li}^{(A, t)}$ are integrated to form a DIR vector $v_{li}^{(t)}$. 
In Section \ref{sec:method:spat_agg}, 
we propose a spatial aggregation mechanism that derives $v_{li}^{(D)}$ and $v_{li}^{(A, t)}$ in a 
bottom-up fashion to 
incrementally fuse information from lower-level industries to upper-level industries. 
Then, in Section \ref{sec:method:temp_agg}, we present a temporal aggregation mechanism that combines these embedding vectors to form a DIR vector.
We summarize DIR and its methodological novelties in Section \ref{sec:method:comp_pic}.

\subsubsection{Spatial Aggregation} \label{sec:method:spat_agg}
We propose a spatial aggregation mechanism to represent 
the assignment-based knowledge about industry ${\mathcal{T}}_{li}$ in period $t$ as a vector $v_{li}^{(A, t)}\in R^{d}$ and embed its definition-based knowledge as a vector $v_{li}^{(D)}\in R^{d}$. We begin by embedding the assignment-based knowledge.

At the leaf level (i.e., $l=L$), $v_{Li}^{(A, t)}$ is randomly initialized and learned as part of the model parameters. However, at level $1 \leq l \leq L-1$, $v_{li}^{(A, t)}$ is derived from the child industries of ${\mathcal{T}}_{li}$:
\begin{equation}\label{eq:spat_agg}
	v_{li}^{(A, t)} = \sum_{k \in {\mathcal{C}}({\mathcal{T}}_{li}) } \beta_{k}^{(li,t)} v_{(l+1)k}^{(A,t)},
\end{equation}
where $0 \le \beta_k^{(li,t)} \le 1$, $\sum_{k \in {\mathcal{C}}({\mathcal{T}}_{li})} \beta_k^{(li,t)} = 1$, $k \in {\mathcal{C}}({\mathcal{T}}_{li}) $ is short for ${\mathcal{T}}_{(l+1)k} \in {\mathcal{C}}({\mathcal{T}}_{li})$, indicating that ${\mathcal{T}}_{(l+1)k}$ is a child industry of ${\mathcal{T}}_{li}$ (similar notation is adopted elsewhere), and $v_{(l+1)k}^{(A,t)}$ denotes the representation of ${\mathcal{T}}_{(l+1)k}$'s assignment-based knowledge in period $t$. The rationale behind Equation \ref{eq:spat_agg} is that the firms assigned to ${\mathcal{T}}_{li}$ are the collection of firms that are assigned to its child industries. Hence, in the equation, we represent ${\mathcal{T}}_{li}$'s assignment-based knowledge as a convex combination of the representations of its child industries' assignment-based knowledge. Clearly, the equation synthesizes assignment-based and structure-based knowledge. It also captures the intuition that child industries are of varying importance through $\beta_k^{(li,t)}$. For example, a child industry abundant in assignment cases might enjoy a larger weight than a child industry with few assignment cases. 
Equation \ref{eq:spat_agg} is applied recursively to levels $l=L-1,L-2,\dots,1$ to represent assignment-based knowledge for every industry at each level. 

Equation \ref{eq:spat_agg} is implemented based on the multi-head self-attention framework \citep{vaswani_attention_2017}. Since the following computations are generic and applicable to any industry in any time period, we simplify the notation by rewriting $v_{(l+1)k}^{(A, t)}$ as $h_k$, dropping the superscript of $\beta_k^{(li, t)}$, and using $\sum_{k}$ to mean $\sum_{k \in {\mathcal{C}}({\mathcal{T}}_{li})}$. As a result, Equation \ref{eq:spat_agg} becomes 
\begin{equation}\label{eq:spat_agg_simp}
	v_{li}^{(A, t)} = \sum_{k} \beta_{k} h_{k}.
\end{equation} 
Next, for expository purposes, we focus on the case of one attention head. 
As the first step, we define
\begin{equation}
	\label{eq:spat_agg_start}
	\tilde{v}_{li}^{(A, t)} = W^{(O)} (\sum_{k} \beta_{k} W^{(V)} h_{k}),
\end{equation}
where $W^{(V)} \in R^{d \times d}$ is the matrix that projects each $h_k$ to a value vector and $W^{(O)} \in R^{d \times d }$ projects the summation to the output vector $\tilde{v}_{li}^{(A, t)} \in R^{d}$.\footnote{When there is more than one attention head, each attention head  produces a single vector $\sum_{k} \beta_{k} W^{(V)} h_{k}$ specific to that attention head.
These vectors are then concatenated into one vector, the length of which equals $hd$, where $h$ is the number of attention heads. In this case, $W^{(O)} \in R^{d \times hd}$ is introduced to project the concatenated vector to the output vector $\tilde{v}_{li}^{(A,t)} \in R^{d}$.} 
The attention weight $\beta_{k}$ should measure the relevance of the assignment-based knowledge of the child industry ${\mathcal{T}}_{(l+1)k}$ to its parent industry ${\mathcal{T}}_{li}$. 
To this end, we need a representation of industry ${\mathcal{T}}_{li}$ that serves as the query vector. 
We do this by reusing the embedding vectors $\{ v_{(l+1)k}^{(A,t)} | \ k \in {\mathcal{C}}({\mathcal{T}}_{li}) \}$, or $\{ h_k \}$ for simplicity, which are available at the time of computing $v_{li}^{(A,t)}$. This design avoids introducing additional parameters for the query vector.
Let $Z$ be the matrix formed by stacking the vectors $\{ h_k \}$ column-wise.
Let $g$ be a function that transforms an input matrix to a vector by taking the mean of each row of the input matrix.\footnote{The choice of function $g$ is flexible as long as it aggregates the vectors of a matrix as one vector. Empirically, we have found that the mean function works well.} Then we use $g(Z)$ as the query representation of ${\mathcal{T}}_{li}$ up to a linear transformation, and define $\beta_k$ as
\begin{equation}
	\label{eq:beta_k}
	\beta_{k} = \frac{\exp\Big( \big(W^{(Q)} g(Z) \big )^T (W^{(K)} h_k) / \sqrt{d} \Big)}{ \sum_{k'} \exp\Big( \big(W^{(Q)} g(Z) \big)^T (W^{(K)} h_{k'}) / \sqrt{d} \Big)},
\end{equation}
where $W^{(Q)} \in R^{d \times d}$ projects $g(Z)$ to a query vector and $W^{(K)} \in R^{d \times d}$ projects each $h_k$ to a key vector. \footnote{The term $\sqrt{d}$ in Equation \ref{eq:beta_k} is used to smooth the distribution of attention weights \citep{vaswani_attention_2017}.}Equation \ref{eq:beta_k} states that $\beta_k$ is measured as the relevance of the piece of knowledge $h_k$ to $g(Z)$, the query representation of parent industry ${\mathcal{T}}_{li}$, up to a linear transformation. 
Finally, to obtain $v_{li}^{(A,t)}$, a post-transformation layer is added on top of $\tilde{v}_{li}^{(A, t)}$ to introduce non-linearity:
\begin{equation}\label{eq:spat_agg_end}
	v_{li}^{(A,t)} = \max \big( W_1 (g(Z) + \tilde{v}_{li}^{(A, t)}) + b_1, 0 \big),
\end{equation}
where $W_1 \in R^{d \times d }$ and $b_1 \in R^{d}$. Equation \ref{eq:spat_agg_end} adds $g(Z)$, the query representation of ${\mathcal{T}}_{li}$, to $\tilde{v}_{li}^{(A,t)}$. This design mimics the residual structure of which the purpose is to facilitate the learning of deep neural networks \citep{he_deep_2016}. The function $\max()$ compares each element of its left argument with zero and returns the greater value. This function is formally called ReLU and is used to introduce non-linearities \citep{goodfellow_deep_2016}.

The computation flow from Equations \ref{eq:spat_agg_start} to \ref{eq:spat_agg_end} is applicable to any industry in any time period and can be generalized as a function
$\textit{MHA}(q, K, V | \Theta)$. The objective of the function is to summarize $n$ pieces of knowledge (e.g., $n$ equals the size of ${\mathcal{C}}({\mathcal{T}}_{li})$ in our case) according to a given query $q$. By applying a linear transformation to $q$, the function produces a query vector. Similarly, by applying linear transformations to the $k$th column vectors of $K$ and $V$, respectively, it generates the key and value vectors for the $k$th piece of knowledge. The query vector is matched against the set of key vectors (e.g., Equation \ref{eq:beta_k}), and the normalized matching scores (e.g., $\beta_{k}$) are used to combine the value vectors as an output vector (e.g., Equation \ref{eq:spat_agg_start}). Lastly, the query $q$ is combined with the output vector, and the summation is transformed non-linearly through Equation \ref{eq:spat_agg_end}.
In the case of spatial aggregation for the assignment-based knowledge of industry ${\mathcal{T}}_{li}$, we have
\begin{equation}\label{eq:spat_agg_simp_assign}
	v_{li}^{(A, t)} = \textit{MHA}(q=g(Z), \ K=Z, \ V=Z|\Theta),
\end{equation}
where the parameters $\Theta$ are defined as
\begin{equation}\label{eq:Theta}
	\Theta=\{ W^{(Q)}, W^{(K)}, W^{(V)}, W_1, b_1, W^{(O)} \}. 
\end{equation}

We have two remarks about the spatial aggregation mechanism defining $v_{li}^{(A, t)}$. First, only the leaf level adds a set of learnable model parameters $\{ v_{Li}^{(A, t)} | i = 1,2,\dots,N_L \}$. The embedding vectors for the assignment-based knowledge at the upper levels are derived recursively from this set of model parameters through Equation \ref{eq:spat_agg_simp_assign}. This design avoids adding a huge set of model parameters for industries beyond the leaf level. 
Second, the set of assignment-based knowledge 
$ \{ {\mathcal{T}}_{li}^{(A,t)} | {\mathcal{T}}_{li} \in {\mathcal{T}} \}$ 
defined by Equation \ref{eq:assign_knowledge} is not used to directly derive the knowledge embedding vectors $ \{ v_{li}^{(A,t)} | {\mathcal{T}}_{li} \in {\mathcal{T}} \}$, but is rather used to indirectly shape them through the learning objective formulated in Section \ref{sec:method:learn}.

To embed the definition-based knowledge ${\mathcal{T}}_{li}^{(D)}$, we employ a DEM that summarizes the semantics of ${\mathcal{T}}_{li}^{(D)}$ as a vector $\bar{v}_{li}^{(D)} \in R^{d}$, i.e.,
\begin{equation}
	\label{eq:ind_def_repr}
	\bar{v}_{li}^{(D)} = \text{DEM}({\mathcal{T}}_{li}^{(D)}).
\end{equation}
The choice of DEM is specified in Section \ref{sec:eval}. Similar to embedding assignment-based knowledge, at the leaf level we set $v_{Li}^{(D)}=\bar{v}_{Li}^{(D)}$, while at level $1\leq l \leq L-1$ we derive $v_{li}^{(D)}$ from $\bar{v}_{li}^{(D)}$ as well as the child industries of ${\mathcal{T}}_{li}$:
\begin{equation}\label{eq:spat_agg_def}
	v_{li}^{(D)} = \gamma_{0}^{(li)} \bar{v}_{li}^{(D)} + \sum_{k \in {\mathcal{C}}({\mathcal{T}}_{li}) } \gamma_{k}^{(li)} v_{(l+1)k}^{(D)},
\end{equation}
where $0 \le \gamma_0^{(li)} \le 1$, $0 \le \gamma_k^{(li)} \le 1$, $\gamma_0^{(li)} + \sum_{k \in {\mathcal{C}}({\mathcal{T}}_{li})} \gamma_k^{(li)} = 1$, $k \in {\mathcal{C}}({\mathcal{T}}_{li}) $ indicates that ${\mathcal{T}}_{(l+1)k}$ is a child industry of ${\mathcal{T}}_{li}$, and $v_{(l+1)k}^{(D)}$ denotes the representation of ${\mathcal{T}}_{(l+1)k}$'s definition-based knowledge.  The intuition of Equation \ref{eq:spat_agg_def} is that business activities covered by any of ${\mathcal{T}}_{li}$'s child industries should also be covered by ${\mathcal{T}}_{li}$. Hence, $v_{li}^{(D)}$ depends on not only the embedded definition-based knowledge of ${\mathcal{T}}_{li}$ but also the definition-based knowledge of its child industries. 
To implement Equation \ref{eq:spat_agg_def}, let $Y$ be the matrix formed by stacking the vectors $ \{ \bar{v}_{li}^{(D)} \} \cup \{  v_{(l+1)k}^{(D)} | k \in {\mathcal{C}}({\mathcal{T}}_{li}) \}$ column-wise. The definition-based knowledge for $1 \leq l \leq L-1$ is then embedded as
\begin{equation}\label{eq:spat_agg_simp_def}
	v_{li}^{(D)} = \textit{MHA}(q=g(Y), \ K=Y, \ V=Y|\Theta),
\end{equation}
where the spatial aggregation parameters $\Theta$ are given in Equation \ref{eq:Theta}.

\subsubsection{Temporal Aggregation}\label{sec:method:temp_agg}
We formulate the DIR vector $v_{li}^{(t)}$ for an industry ${\mathcal{T}}_{li}$ as a convex combination of vectors $v_{li}^{(D)}, v_{li}^{(A, 1)}, v_{li}^{(A, 2)}, \dots, v_{li}^{(A, t)} $, which represent  ${\mathcal{T}}_{li}$'s definition-based knowledge and assignment-based knowledge up to period $t$. Accordingly, we have
\begin{equation}\label{eq:temp_agg_basic}
	v_{li}^{(t)} = \alpha_0^{(li,t)} v_{li}^{(D)} + \alpha_1^{(li,t)} v_{li}^{(A, 1)} + \alpha_2^{(li,t)} v_{li}^{(A, 2)} + \dots + \alpha_{t}^{(li,t)} v_{li}^{(A, t)},
\end{equation}
where $0 \le \alpha_k^{(li,t)} \le 1$ and $\sum_{k=0}^{t} \alpha_k^{(li,t)} = 1$. The scalar $\alpha_k^{(li,t)}$ serves as the attention weight placed on the $k$th piece of knowledge and is specific to industry ${\mathcal{T}}_{li}$ in period $t$. Intuitively, if a piece of knowledge is useful for classifying a firm, it should have a relatively large weight. Attention weights vary across industries and time for the following reasons. First, some industries might be more informatively defined than others by using more specific words conveying their covered business activities. As a result, definition-based knowledge of different industries is not equally important for industry assignment. Second, attention weights are time-specific, because the importance of pieces of knowledge should be evaluated within the knowledge set that is available by the time period considered and reevaluated when new pieces of knowledge are introduced in subsequent periods.

We implement Equation \ref{eq:temp_agg_basic} in a similar way to the spatial aggregation mechanism.
Using the notation $\textit{MHA}(q,K,V)$ developed in Section \ref{sec:method:spat_agg}, we define
\begin{equation}\label{eq:temp_agg_simp}
	v_{li}^{(t)} = \textit{MHA}(q=v_{li}^{(A, t)}, \ K=M, \ V=M \ | \ \Theta'),
\end{equation}
where $M$ is the matrix formed by stacking the vectors $v_{li}^{(D)}, v_{li}^{(A, 1)}, v_{li}^{(A, 2)}, \dots, v_{li}^{(A, t)}$ column-wise. 
We use $v_{li}^{(A,t)}$ in Equation \ref{eq:temp_agg_simp} as the query vector. Because the assignment-based knowledge in period $t$ represents the most up-to-date expert knowledge about which firms should be classified into industry ${\mathcal{T}}_{li}$, measuring the attention weights based on this knowledge generates an integrated knowledge representation that is best suited for classifying a firm in period $t$.
The computation behind Equation \ref{eq:temp_agg_simp} contains two steps. First, a vector $\tilde{v}_{li}^{(t)}$ is computed in a similar way to Equation \ref{eq:spat_agg_start} by averaging knowledge embedding vectors $v_{li}^{(D)}, v_{li}^{(A, 1)}, v_{li}^{(A, 2)}, \dots, v_{li}^{(A, t)}$ (which are column vectors of $M$) based on attention weights for query $v_{li}^{(A, t)}$. Second, a post-transformation layer analogous to Equation \ref{eq:spat_agg_end} is imposed, which is defined as $v_{li}^{(t)} = \max \big( W_1' (v_{li}^{(A,t)} + \tilde{v}_{li}^{(t)}) + b_1', 0 \big)$ where $v_{li}^{(A,t)}$ is added to $\tilde{v}_{li}^{(t)}$ to facilitate the learning of our model.

The set of \textit{MHA} parameters in this case is denoted by $\Theta'$ and has a similar structure to $\Theta$ (Equation \ref{eq:Theta}), but is parameterized independently. Formally, $\Theta'$ is given by 
\begin{equation}\label{eq:ThetaPrime}
	\Theta'=\{ W'^{(Q)}, W'^{(K)}, W'^{(V)}, W'_1, b'_1, W'^{(O)}\}, \
\end{equation}
and these parameters are shared across industries and time periods.

\subsubsection{Summary} \label{sec:method:comp_pic}

Dynamic industry representation distinguishes our method from existing industry assignment methods through its novel representation of an industry as a sequence of time-specific vectors that are derived by integrating definition-based, assignment-based, and structure-based knowledge through the proposed temporal and spatial aggregation mechanisms. Although these mechanisms are built upon the multi-head self-attention framework, applying the framework to our problem requires defining and solving problem-specific details. In this regard, the novelty of the temporal and spatial aggregation mechanisms lies in the formulation of the three embedding vectors in Equations \ref{eq:temp_agg_basic}, \ref{eq:spat_agg}, and \ref{eq:spat_agg_def}, as well as the specification of queries, keys, and values in Equations \ref{eq:temp_agg_simp}, \ref{eq:spat_agg_simp_assign}, \ref{eq:spat_agg_simp_def}, 
which are designed to leverage the three types of expert knowledge for dynamic industry representation. 

\subsection{Hierarchical Assignment} \label{sec:method:ha}

Given firm representation $x^{(t,j)}$ (Equation \ref{eq:firm_repr}) and dynamic industry representation $v_{li}^{(t)}$ (Equation \ref{eq:temp_agg_basic}), we define the compatibility score between firm $j$ and industry ${\mathcal{T}}_{li}$ in period $t$ as
\begin{equation}
	\label{eq:compat_score}
	s({\mathcal{T}}_{li}, j , t ) = \exp({v_{li}^{(t)}}^T x^{(t, j)}),
\end{equation}
which is the exponential of the inner product of the firm representation and the industry representation. In deep learning, it is common to measure the compatibility between two vector representations using their inner product followed by an exponential transformation \citep{goodfellow_deep_2016}. In our context, the larger the compatibility score, the better the business activities of the firm fit into the scope of the industry. 

Let $P({\mathcal{T}}_{li}| {\mathcal{P}}({\mathcal{T}}_{li}), j, t)$ be the probability that in period $t$, firm $j$ is assigned to industry ${\mathcal{T}}_{li}$ among all the industries at level $l$, given that it has already been assigned to the parent industry ${\mathcal{P}}({\mathcal{T}}_{li})$ of ${\mathcal{T}}_{li}$. To measure $P({\mathcal{T}}_{li}| {\mathcal{P}}({\mathcal{T}}_{li}), j, t)$, we observe the following hierarchy constraint for ICSs: 

\begin{definition}[\textbf{The Hierarchy Constraint}]
	\label{def:hier_constraint}
	If a firm is assigned to an industry $z$, it must also belong to one of industry $z$'s child industries. Likewise, if a firm is assigned to an industry $y$, it should also belong to $y$'s parent industry.
\end{definition}

Consider the ICS in Figure \ref{fig:ic_tree} as an example. By the hierarchy constraint, a firm assigned to industry ${\mathcal{T}}_{11}$ must also belong to either ${\mathcal{T}}_{21}$ or ${\mathcal{T}}_{22}$. Similarly, a firm assigned to industry ${\mathcal{T}}_{23}$ should also belong to ${\mathcal{T}}_{12}$.  

By the hierarchy constraint, if a firm has been assigned to industry ${\mathcal{P}}({\mathcal{T}}_{li})$ (i.e., the parent industry of ${\mathcal{T}}_{li}$) at level $l+1$, it must belong to one of the child industries of ${\mathcal{P}}({\mathcal{T}}_{li})$ at level $l$, i.e., ${\mathcal{C}}({\mathcal{P}}({\mathcal{T}}_{li}))$. Therefore, given that a firm has been assigned to industry ${\mathcal{P}}({\mathcal{T}}_{li})$ at level $l+1$, the search space for the firm's industry assignment at level $l$ reduces from all of the industries at that level to ${\mathcal{C}}({\mathcal{P}}({\mathcal{T}}_{li}))$. Accordingly, we can formulate $P({\mathcal{T}}_{li}| {\mathcal{P}}({\mathcal{T}}_{li}), j, t)$ as the compatibility score $s({\mathcal{T}}_{li}, j, t)$ normalized within ${\mathcal{C}}({\mathcal{P}}({\mathcal{T}}_{li}))$:
\begin{equation}
	\label{eq:relative_prob}
	P({\mathcal{T}}_{li}| {\mathcal{P}}({\mathcal{T}}_{li}), j, t)  
	=\frac{s({\mathcal{T}}_{li}, j, t)}{\sum_{i' \in {\mathcal{C}} \big({\mathcal{P}}({\mathcal{T}}_{li})\big) } s({\mathcal{T}}_{li'}, j, t)}, 
\end{equation}
where $s({\mathcal{T}}_{li}, j, t)$ can be computed using Equation \ref{eq:compat_score}. We illustrate the computation of $P({\mathcal{T}}_{li}| {\mathcal{P}}({\mathcal{T}}_{li}), j, t)$ with the following example. 

\begin{example}
	Figure \ref{fig:compat_scores} gives an example of compatibility scores between firm $j$ and each industry in the ICS illustrated in Figure \ref{fig:ic_tree}. Figure \ref{fig:assign_prob_rel} shows the probabilities $P({\mathcal{T}}_{li}| {\mathcal{P}}({\mathcal{T}}_{li}), j, t)$ computed from these compatibility scores using Equation \ref{eq:relative_prob}. Since ${\mathcal{T}}_{01}$ denotes the entire ICS and firm $j$ must belong to an industry in the ICS, we have $P({\mathcal{T}}_{01}|j, t)=1$. Consider the computation of $P({\mathcal{T}}_{11}| {\mathcal{T}}_{01}, j, t)$. Examining the ICS given in Figure \ref{fig:ic_tree}, we note that ${\mathcal{C}}({\mathcal{T}}_{01}) = \{{\mathcal{T}}_{11},{\mathcal{T}}_{12}\}$ and ${\mathcal{P}}({\mathcal{T}}_{11}) = \{{\mathcal{T}}_{01}\}$. By Equation \ref{eq:relative_prob}, we have
	$$P({\mathcal{T}}_{11}| {\mathcal{T}}_{01}, j, t)  
	=\frac{s({\mathcal{T}}_{11}, j, t)}{\sum_{i' \in {\mathcal{C}} ({\mathcal{T}}_{01}) } s({\mathcal{T}}_{1i'}, j, t)}=\frac{s({\mathcal{T}}_{11}, j, t)}{s({\mathcal{T}}_{11}, j, t)+s({\mathcal{T}}_{12}, j, t)}=0.4.$$

	\begin{figure}[h]
		\centering
		\begin{tikzpicture}
			\tikzstyle{level 1}=[sibling distance=7.5cm, level distance=1cm]
			\tikzstyle{level 2}=[sibling distance=3cm, level distance=1cm]
			\node (root) {\small${\mathcal{T}}_{01}$}
			child {
				node (11) {\small$s({\mathcal{T}}_{11},j, t)=4$}
				child {
					node (21) {\small$s({\mathcal{T}}_{21},j, t)=2.5$}
				}
				child {
					node (22) {\small$s({\mathcal{T}}_{22},j, t)=2.5$}
				}
			}
			child {
				node (12) {\small$s({\mathcal{T}}_{12},j, t)=6$}
				edge from parent [style={solid}]
				child {
					node (23) {\small$s({\mathcal{T}}_{23},j, t)=1$}
					edge from parent [style={solid}]
				}
				child {
					node (24) {\small$s({\mathcal{T}}_{24},j, t)=1$}
					edge from parent [style={solid}]
				}
				child {
					node (25) {\small$s({\mathcal{T}}_{25},j, t)=8$}
					edge from parent [style={solid}]
				}
			};
		\end{tikzpicture}
		\caption{Compatibility Scores between Firm $j$ and Each Industry in the Example ICS in Figure \ref{fig:ic_tree}}
		\label{fig:compat_scores}
	\end{figure}
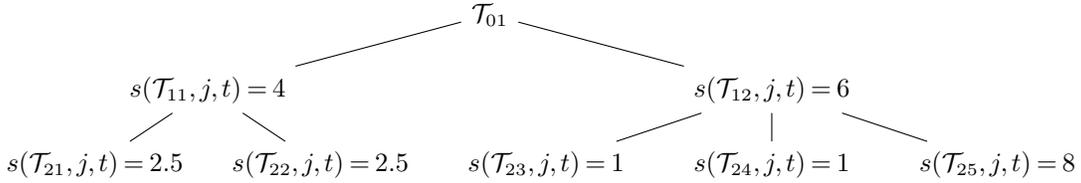
	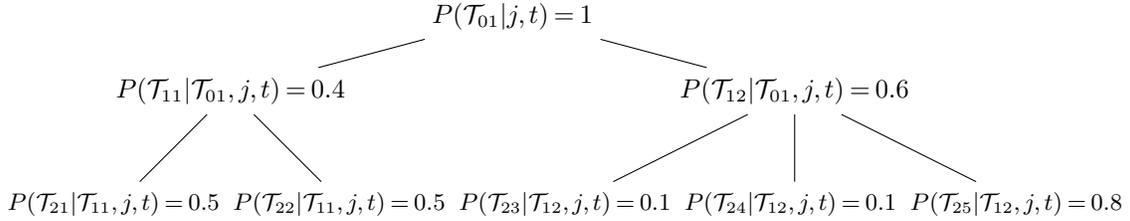
\begin{figure}[h]
		\centering
		\begin{tikzpicture}
			\tikzstyle{level 1}=[sibling distance=7.5cm, level distance=1cm]
			\tikzstyle{level 2}=[sibling distance=3cm, level distance=1.5cm]
			\node (root) {\small$P({\mathcal{T}}_{01}|j, t)=1$}
			child {
				node (11) {\small$P({\mathcal{T}}_{11}|{\mathcal{T}}_{01}, j, t)=0.4$}
				child {
					node[align=center] (21) {\footnotesize$P({\mathcal{T}}_{21}|{\mathcal{T}}_{11}, j, t)=0.5$
					} 
				}
				child {
					node[align=center] (22) {\footnotesize$P({\mathcal{T}}_{22}|{\mathcal{T}}_{11}, j, t)=0.5$
					}
				}
			}
			child {
				node (12) {\small$P({\mathcal{T}}_{12}|{\mathcal{T}}_{01}, j, t)=0.6$}
				edge from parent [style={solid}]
				child {
					node[align=center] (23) {\footnotesize$P({\mathcal{T}}_{23}|{\mathcal{T}}_{12}, j, t)=0.1$
					}
					edge from parent [style={solid}]
				}
				child {
					node[align=center] (24) {\footnotesize$P({\mathcal{T}}_{24}|{\mathcal{T}}_{12}, j, t)=0.1$
					}
					edge from parent [style={solid}]
				}
				child {
					node[align=center] (25) {\footnotesize$P({\mathcal{T}}_{25}|{\mathcal{T}}_{12}, j, t)=0.8$
					}
					edge from parent [style={solid}]
				}
			};
		\end{tikzpicture}
		\caption{Probability $P({\mathcal{T}}_{li}| {\mathcal{P}}({\mathcal{T}}_{li}), j)$ for Each Industry in the Example ICS}
		\label{fig:assign_prob_rel}
	\end{figure}
\end{example}

\noindent The objective of hierarchical assignment is to compute $P({\mathcal{T}}_{li}| j, t)$, the probability that in period $t$, firm $j$ is assigned to industry ${\mathcal{T}}_{li}$ among all the industries at level $l$. In accordance with Bayes' theorem, we have
\begin{equation}
	\label{eq:bayesfact}
	P( {\mathcal{T}}_{li} | {\mathcal{P}}({\mathcal{T}}_{li}), j, t) = \frac{P( {\mathcal{P}}({\mathcal{T}}_{li})| {\mathcal{T}}_{li}, j, t) P( {\mathcal{T}}_{li} | j, t)  }{P( {\mathcal{P}}({\mathcal{T}}_{li}) | j, t)}.
\end{equation}
According to the hierarchy constraint, if firm $j$ is assigned to industry ${\mathcal{T}}_{li}$, the firm must belong to ${\mathcal{T}}_{li}$'s parent industry $P( {\mathcal{T}}_{li})$ as well. Hence, we have $P( {\mathcal{P}}({\mathcal{T}}_{li})| {\mathcal{T}}_{li}, j, t)=1$. Accordingly, Equation \ref{eq:bayesfact} can be rewritten as
\begin{equation}
	\label{eq:bayesfact2}
	P( {\mathcal{T}}_{li} | j, t) = P( {\mathcal{T}}_{li} | {\mathcal{P}}({\mathcal{T}}_{li}), j, t) P( {\mathcal{P}}({\mathcal{T}}_{li}) | j, t).
\end{equation}
By applying Equation \ref{eq:bayesfact2} to the last term in Equation \ref{eq:bayesfact2}, we have 
\begin{equation*}
	P( {\mathcal{P}}({\mathcal{T}}_{li}) | j, t) = P( {\mathcal{P}}({\mathcal{T}}_{li}) | {\mathcal{P}}^{2}({\mathcal{T}}_{li}), j, t) P( {\mathcal{P}}^{2}({\mathcal{T}}_{li}) | j, t).
\end{equation*}
Thus, $P( {\mathcal{T}}_{li} | j, t)$ can be expanded recursively to the root level of the ICS: 
\begin{equation*}
	\begin{split}
		P( {\mathcal{T}}_{li} | j, t) &= P( {\mathcal{T}}_{li} | {\mathcal{P}}({\mathcal{T}}_{li}), j, t) P( {\mathcal{P}}({\mathcal{T}}_{li}) | {\mathcal{P}}^{2}({\mathcal{T}}_{li}), j, t) P( {\mathcal{P}}^2({\mathcal{T}}_{li}) | j, t) \\
		&= \dots \\
		&= \Big( \Pi_{m=0}^{l-1} P( {\mathcal{P}}^{m}({\mathcal{T}}_{li}) | {\mathcal{P}}^{m+1}({\mathcal{T}}_{li}), j, t) \Big) P({\mathcal{T}}_{01}|j, t). 
	\end{split}
\end{equation*}
Recall that $P({\mathcal{T}}_{01}|j, t)=1$, because ${\mathcal{T}}_{01}$ denotes the entire ICS. Therefore, $P( {\mathcal{T}}_{li} | j, t)$ is factorized as
\begin{equation} \label{eq:p_abs_upwards}
	P( {\mathcal{T}}_{li} | j, t) = \Pi_{m=0}^{l-1} P( {\mathcal{P}}^{m}({\mathcal{T}}_{li}) | {\mathcal{P}}^{m+1}({\mathcal{T}}_{li}), j, t), 
\end{equation}
each factor of which can be computed using Equation \ref{eq:relative_prob}.

\begin{example}\label{ex:ha_numeric}
	Continuation of Example 3. By applying Equation \ref{eq:p_abs_upwards}, we can compute $P({\mathcal{T}}_{25}|j, t)$ as
	$$P({\mathcal{T}}_{25}|j, t)= P({\mathcal{T}}_{25}|{\mathcal{T}}_{12},j, t)P({\mathcal{T}}_{12}|{\mathcal{T}}_{01},j, t)=0.8 \times 0.6=0.48.$$
\end{example}

Hierarchical assignment is distinct from the flat assignment employed by existing industry assignment methods, where a flattened class space is constructed with each class corresponding to an industry at the focal level. Flat assignment only considers industries at the focal level and ignores industries that are above them, thereby neglecting structure-based knowledge. In contrast, hierarchical assignment considers industries across hierarchical levels and incorporates structure-based knowledge into the factorization structure for computing $P( {\mathcal{T}}_{li} | j, t)$ (i.e., Equation \ref{eq:p_abs_upwards}).

\subsection{DeepIA}\label{sec:method:learn}

DeepIA is trained with past firm--industry assignments ${\mathcal{D}}=\{ (j, y^{(t, j)}) \ | \ j \in {\mathcal{U}}, \ y^{(t, j)} \in {\mathcal{T}}_{l^{*}}, \ t \in 1:T \}$. For each $(j, y^{(t, j)}) \in {\mathcal{D}}$, the representation of firm $j$ in period $t$ is derived using Equation \ref{eq:firm_repr}. Next, the industry representations in periods $1:T$ are derived by integrating definition-based, assignment-based, and structured-based knowledge through the temporal and spatial aggregation mechanisms. Finally, for each $(j, y^{(t, j)}) \in {\mathcal{D}}$, the probability $P(y^{(t,j)}|j)$ that firm $j$ belongs to industry $y^{(t, j)}$ in period $t$ is computed using Equation \ref{eq:p_abs_upwards}. Let $\Phi$ denote the model parameters of DeepIA,  we have 
\begin{equation}
	\label{eq:all_params}
	\Phi =\{\Theta,\Theta',\Delta,\Gamma\}, 
\end{equation}
where the parameters $\Theta$ and $\Theta'$ are respectively specified in Equations \ref{eq:Theta} and \ref{eq:ThetaPrime}, $\Delta$ denotes the parameters associated with transformation layer $e_f$ in Equation \ref{eq:firm_repr}, and $\Gamma$ contains randomly initialized representations of leaf-level assignment-based knowledge (see Section \ref{sec:method:spat_agg})---specifically, $\Gamma=\{ v_{Li}^{(A,t)} \ | \ i\in {\mathcal{T}}_L,  t=1:T \}$.\footnote{We compute the model complexity of DeepIA based on Equation \ref{eq:all_params}. Specifically, $\Theta$ is of size 
	$4hd^2+d^2+d=(4h+1)d^2+d$, where $4hd^2$ accounts for the number of parameters of the $h$ attention heads each with four $d \times d$ matrices (i.e., $ W^{(Q)}, W^{(K)}, W^{(V)}$, $W^{(O)}$), $d^2$ is the size of the $d \times d$ matrix $W_1$, and $d$ is the length of vector $b_1$. 
	The parameter set $\Theta'$  has the same structure as $\Theta$ and therefore is of size $(4h+1)d^2+d$. 
	The transformation layer $e_f$ will be later specified in Section \ref{sec:eval:bench} as a multi-layer perceptron with two layers, and hence its parameter set $\Delta$ is of size $(2d\times d+2d)+( d\times 2d+d)=4d^2+3d$.
	Lastly, $\Gamma$ is of size $N_L T d$, where $N_L$ is the number of industries at leaf level $L$ and $d$ is the size of embedding vectors. 
	In conclusion, $\Phi$, the parameter set of DeepIA, is of size $(8h+6)d^2 + (N_L T+5)d$. \label{fn:DeepIA_paramsize}}

The model parameters for DeepIA are learned by optimizing the following objective function through mini-batch gradient descent:
\begin{equation}
	\label{eq:optimization}
	\Phi^* = \argmin_{\Phi} \sum_{(j, y^{(t,j)}) \in {\mathcal{D}}} -\log P(y^{(t,j)}|j,t). 
\end{equation}

\begin{algorithm}[h]
	\caption{The Training Procedure for DeepIA}
	\label{alg:learn}
	\textbf{Input:} Firms' business description documents, focal level industry $l^*$, past firm--industry assignments ${\mathcal{D}}$, encoded expert knowledge $({\mathcal{T}}_{li}^{(D)}, {\mathcal{T}}_{li}^{(S)}, {\mathcal{T}}_{li}^{(A,t)})$ for all ${\mathcal{T}}_{li} \in {\mathcal{T}}$ and $t \in 1:T$\\
	\textbf{Output:} Learned model parameters $\Phi^*$
	\begin{algorithmic}[1]
		\State Pretrain a DEM on firms' business description documents \label{alg:learn:l:pt1}
		\State Pretrain a DEM on $\{{\mathcal{T}}_{li}^{(D)} | {\mathcal{T}}_{li} \in {\mathcal{T}} \}$
		\label{alg:learn:l:pt2}
		\State Initialize $\Phi$ \Comment{Eq. \ref{eq:all_params}} \label{alg:learn:l:init}
		\For {epoch $\in 1:N_{\text{epoch}}$} \Comment{ $N_{\text{epoch}}$: number of epochs} \label{alg:learn:l:ep1}
		\State Permute ${\mathcal{D}}$ and partition it into batches ${\mathcal{B}}_1 \cup {\mathcal{B}}_2 \cup \dots \cup  {\mathcal{B}}_{N_{\text{batch}}}$ \Comment{ $N_{\text{batch}}$: number of batches} \label{alg:learn:l:part}
		\For {$k \in 1:N_{\text{batch}}$} \label{alg:learn:l:b1}
		\For {$l \in L-1:1$} \label{alg:learn:l:dir1}
		\State Compute $\{ v_{li}^{(A,t)} \ | \ i \in {\mathcal{T}}_{l}, t \in 1:T  \}$ \Comment{Eq. \ref{eq:spat_agg_simp_assign}}
		\State Compute $\{ v_{li}^{(D)} \ | \ i \in {\mathcal{T}}_{l} \}$ \Comment{Eq. \ref{eq:spat_agg_simp_def}}
		\EndFor 
		\For {${\mathcal{T}}_{li} \in {\mathcal{T}}$}
		\State Compute $ \{ v_{li}^{(t)} \ | \ t \in 1:T \}$ \Comment{Eq. \ref{eq:temp_agg_simp}}
		\EndFor \label{alg:learn:l:dir2}
		\For {$(j, y^{(t, j)}) \in {\mathcal{B}}_k$} \label{alg:learn:l:ha1}
		\State Compute $ x^{(t,j)}$ \Comment{Eq. \ref{eq:firm_repr}}
		\For {${\mathcal{T}}_{li}$ in ${\mathcal{T}}$}
		\State Compute $ s({\mathcal{T}}_{li}, j , t )$ \Comment{Eq. \ref{eq:compat_score}}   
		\State Compute $P({\mathcal{T}}_{li} |  {\mathcal{P}}({\mathcal{T}}_{li}), j, t)$ \Comment{Eq. \ref{eq:relative_prob}}
		\EndFor
		\For {${\mathcal{T}}_{l^*i}$ in ${\mathcal{T}}_{l^*}$}
		\State Compute $P({\mathcal{T}}_{l^*i} | j, t)$ \Comment{Eq. \ref{eq:p_abs_upwards}}
		\EndFor \label{alg:learn:l:ha2}
		\EndFor
		\State Compute loss ${\mathcal{L}} = \sum_{(j, y^{(t,j)}) \in {\mathcal{B}}_k} -\log P(y^{(t,j)}|j, t)$ \label{alg:learn:l:loss}
		\State Compute gradients $\partial {\mathcal{L}} / \partial \Phi$
		\State Update $\Phi$ with gradient descent
		\EndFor \label{alg:learn:l:b2}
		\EndFor
		\State \Return $\Phi$ as $\Phi^*$
	\end{algorithmic} 
\end{algorithm}
\noindent The training procedure for finding $\Phi^*$ is summarized in Algorithm \ref{alg:learn}. 
In lines \ref{alg:learn:l:pt1} and \ref{alg:learn:l:pt2} of the algorithm, two DEMs are respectively trained on the corpus of firms' business description documents and the corpus of industry definitions. 
Line \ref{alg:learn:l:init} randomly initializes all of the model parameters. Starting from line \ref{alg:learn:l:ep1}, the algorithm is executed for $N_\text{epoch}$ epochs. At the beginning of each epoch, the assignment cases in ${\mathcal{D}}$ are randomly permuted and then partitioned into $N_\text{batch}$ batches. Each batch ${\mathcal{B}}_k$ consists of a subset of the assignment cases from ${\mathcal{D}}$, with the size of the batch specified in Section \ref{sec:eval:bench}. The computation steps between line \ref{alg:learn:l:b1} and line \ref{alg:learn:l:b2} illustrate how the model parameters can be iteratively learned in a batch-by-batch style. At the beginning of each batch, all of the DIR vectors (Section \ref{sec:method:dir}) are prepared between lines \ref{alg:learn:l:dir1} and \ref{alg:learn:l:dir2}. Then the HA procedure (Section \ref{sec:method:ha}) is performed between lines \ref{alg:learn:l:ha1} and \ref{alg:learn:l:ha2} for each assignment case in the batch. Next, the loss of the batch to be minimized is computed at line \ref{alg:learn:l:loss}, with the gradients derived in the next line. Lastly, based on the gradient information, the model parameters are updated through gradient descent. 

\begin{algorithm}[h!]
	\caption{The Inference Procedure of DeepIA}
	\textbf{Input:} Business description document $doc^{(T+1,j)}$ of firm $j$ in period $T+1$ and focal-level industry $l^*$ \\
	\textbf{Output:}  Predicted industry assignment $\hat{y}^{(T+1,j)}$
	\label{alg:infer}
	\begin{algorithmic}[1]
		\State Compute $\text{DEM}(doc^{(T+1,j)})$ and then $x^{(T+1,j)}$ \Comment{Eq. \ref{eq:firm_repr}} \label{alg:infer:l:firm_repr}
		\For {${\mathcal{T}}_{li}$ in ${\mathcal{T}}$} \label{alg:infer:l:ha1}
		\State Compute $ s({\mathcal{T}}_{li}, j, T+1)$ \Comment{Eq. \ref{eq:compat_score}}   
		\State Compute $P({\mathcal{T}}_{li} |  {\mathcal{P}}({\mathcal{T}}_{li}), j, T+1)$ \Comment{Eq. \ref{eq:relative_prob}}
		\EndFor
		\For {${\mathcal{T}}_{l^*i}$ in ${\mathcal{T}}_{l^*}$}
		\State Compute $P({\mathcal{T}}_{l^*i} |j, T+1)$ \Comment{Eq. \ref{eq:p_abs_upwards}}
		\EndFor \label{alg:infer:l:ha2}
		\State $\hat{y}^{(T+1,j)}=\arg\max_{{\mathcal{T}}_{l^*i} \in {\mathcal{T}}_{l^*}} P({\mathcal{T}}_{l^*i} |j, T+1)$
		\State \Return $\hat{y}^{(T+1,j)}$
	\end{algorithmic} 
\end{algorithm}

Once trained, DeepIA can be applied to classify an unassigned firm $j \in {\mathcal{U}}$ in period $T+1$ into a focal-level industry following the steps in Algorithm \ref{alg:infer}. Specifically, the representation of the unassigned firm is derived at line \ref{alg:infer:l:firm_repr} based on its most up-to-date business description document. Next, the assignment probabilities are computed between lines \ref{alg:infer:l:ha1} and \ref{alg:infer:l:ha2}. Note that the computation of $s({\mathcal{T}}_{li}, j , T+1)$ requires $v_{li}^{(A, T+1)}$, which is not available because the true industry assignment of firm $j$ in period $T+1$ is unknown at the time of prediction. Therefore, we use $s({\mathcal{T}}_{li}, j , T)$ as a surrogate for $s({\mathcal{T}}_{li}, j , T+1)$.
In the last line, the industry assigned to the firm, $\hat{y}^{(T+1,j)}$, is the one with the highest assignment probability at the focal industry level.

\section{Empirical Evaluation} \label{sec:eval}

We benchmark DeepIA against several prevalent methods on the tasks of assigning firms to industries of two widely used ICSs: North American Industry Classification System (NAICS) and Global Industry Classification Standard (GICS). We report evaluation results with NAICS in this section. Similar evaluation results are obtained using GICS and reported in Appendix \ref{ap:gics} for space consideration.

\subsection{Data and Evaluation Procedure} \label{sec:eval:data}

Our evaluation was conducted using public data. Specifically, we acquired a firm universe from the Compustat Company Header History (COMPHIST) data provided by Wharton Research Data Services (WRDS). Each row in COMPHIST records the information of a firm in a particular time period. We focused on the following fields in COMPHIST: GVKEY of a firm, which uniquely identifies a firm in the database, HNAICS of a firm, which gives  six-digit NAICS code assigned to a firm, as well as the time period within which the information is valid. For example, one record for firm \textit{Costco Wholesale Corporation} is (GVKEY=29028, HNAICS=452910) and this record is valid in year 2012. That is, \textit{Costco Wholesale Corporation} is assigned to NAICS industry \textit{Warehouse Clubs and Supercenters} (coded as 452910) in year 2012. 
From COMPHIST, we collected a dataset of firm-industry assignments from year 2012 to year 2016. Each record of the dataset shows the assignment of a firm to an industry in a year. Table \ref{tb:stat} reports the summary statistics of the dataset. In this table, column $n$ indicates the number of firm-industry assignments in a year. In the dataset, a firm is only assigned to one industry in a year; hence, $n$ also shows the number of unique firms in a year. For instance, there are 5,322 firm-industry assignments or unique firms in year 2012.  Column $n_{\text{new}}$ means the number of new firms in a year. A firm is considered new in a year if it does not appear in any previous year(s). For example, out of 5,402 firms in year 2013, 692 are new firms.

\begin{table}[h!]
	\caption{Summary Statistics of the Firm-Industry Assignment Dataset}
	\centering
	\begin{tabular}{lcc}
		\hline
		Year & $n$ & $n_{\text{new}}$ \\ \hline
		2012 & 5,322 & - \\ 
		2013 & 5,402 & 692 \\ 
		2014 & 5,360 & 421 \\ 
		2015 & 5,335 & 434 \\ 
		2016 & 5,054 & 258 \\ \hline
	\end{tabular}
	\label{tb:stat}
\end{table}

\begin{figure}[h]
	\centering
	\includegraphics[width=0.5\textwidth]{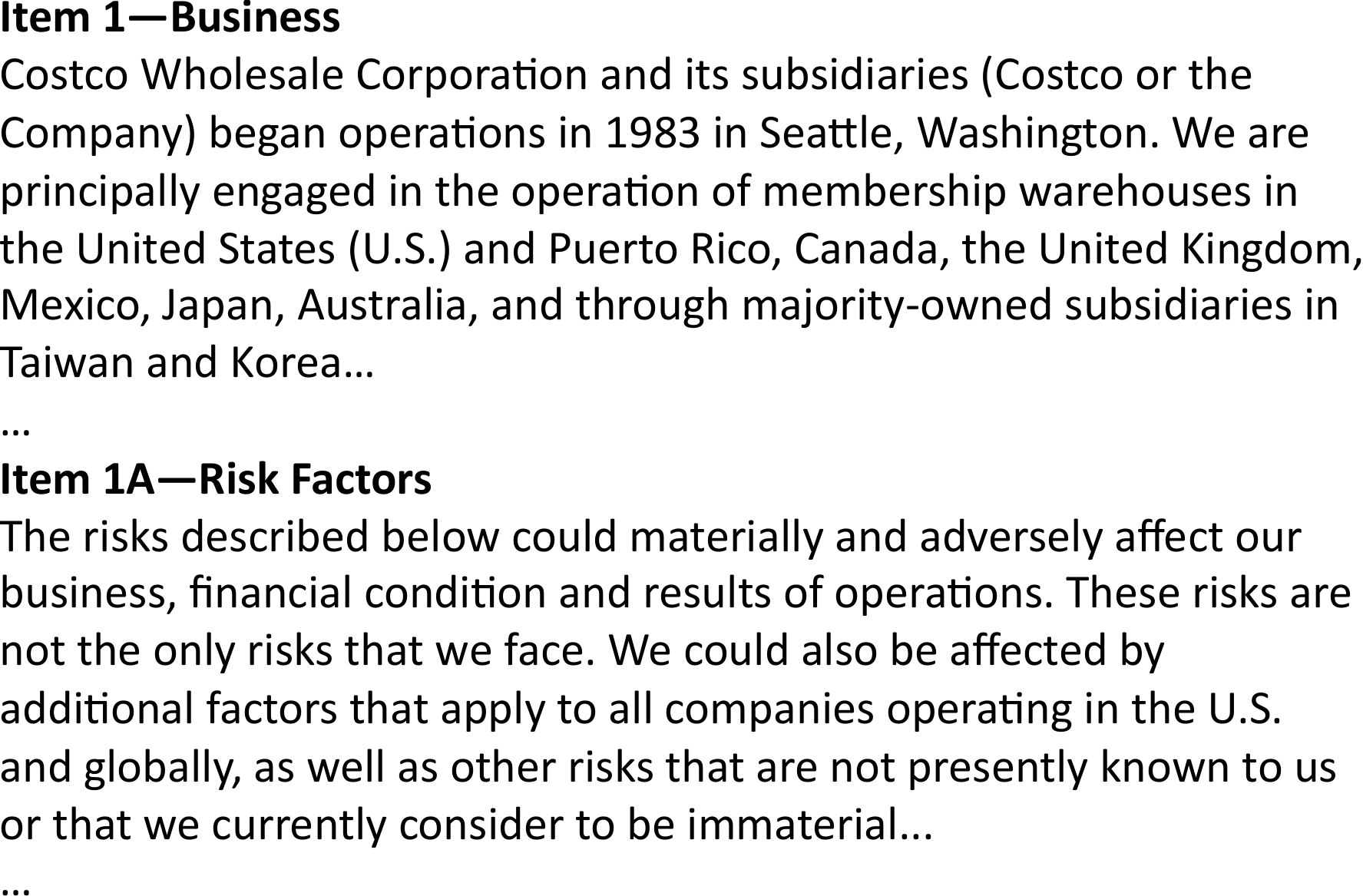}
	\caption{An Excerpt of Costco's Business Description Document in Year 2012}\label{fig:10k}
\end{figure}

For each firm-year observation in the firm-industry assignment dataset, we constructed the business description document of the firm as Items 1 and 1A of the firm's 10-K report filed in that year. We employ 10-K reports as the source of business description because Items 1 and 1A of a firm's 10-K report contain accurate, up-to-date, and rich information about its business activities and associated risk factors \citep{hoberg_text-based_2016}. Specifically, we collected business description documents from the Stage One 10-X Parse
Data, which contain 10-K reports preprocessed by \cite{loughran_textual_2016}.\footnote{The data set can be downloaded at \url{https://sraf.nd.edu/data/stage-one-10-x-parse-data/}.}  On average, a business description document in our evaluation consists of 10,000 words. As an example, Figure \ref{fig:10k} shows an excerpt of the business description document of \textit{Costco Wholesale Corporation} in year 2012. 

The structure of NAICS as well as the definitions of its industries are revised every five years and the revision occurred in year 2012 (NAICS 2012) covers year 2012 to year 2016. Therefore, we used NAICS 2012 in our evaluation. Table \ref{tb:n_ind} summarizes the structure of NAICS 2012, where $l$ denotes industry level, $N_l$ indicates the number of industries at level $l$, and $D_l$ means the number of digits for an industry code at level $l$.\footnote{The official website for NAICS 2012 can be accessed at \url{https://www.census.gov/naics/?58967?yearbck=2012}, which contains its detailed structural information and definitions of its industries.} For example, there are 1,065 unique industries at level 5 of NAICS 2012 and each industry at this level is denoted by a six-digit code.

\begin{table}[h]
	\caption{Structure of NAICS 2012}
	\centering
	\begin{tabular}{lccccc}
		\hline
		$l$  & 1 & 2 & 3 & 4 & 5 \\ \hline
		$N_l$ & 20 & 99 & 312 & 713 & 1,065 \\ 
		$D_l$ & 2 & 3 & 4 & 5 & 6 \\ \hline
	\end{tabular}
	\label{tb:n_ind}
\end{table}

Having introduced data, we detail our evaluation procedure. Each method (ours or benchmark) takes firm-industry assignment data and business description documents from year 2012 to year $T$ as training data to predict industry assignment for each new firm in year $T+1$ according to NAICS 2012. We focus on predicting industry assignments for new firms because there are no assignment records for these firms in the training data. To fine-tune hyperparameters of each method, we use assignment data of new firms in year $T$ as validation data. Take $T=2015$ as an example. Training data for this evaluation contain firm-industry assignments from year 2012 to year 2015 as well as business description document for each firm-year observation in the same time period. Validation data consist of industry assignments of new firms in year 2015. The objective of the evaluation is to predict industry assignment for each new firm in year 2016. The performance of a method is evaluated by comparing its predicted industry assignments against true industry assignments in year $T+1$ based on commonly used metrics: accuracy and macro-F1 \citep{narasimhan_optimizing_2016, wood_automated_2017}. Specifically, accuracy is the percentage of correctly classified firms. Let $\text{TP}_i$ be the number of firms that are predicted belonging to industry ${\mathcal{T}}_{l^*i}$ and actually belong to the industry, $i=1,2,\dots,N_{l^*}$, where $l^*$ denotes the focal level and $N_{l^*}$ is the number of industries at the level. Similarly, let $\text{FP}_i$ be the number of firms that are predicted belonging to industry ${\mathcal{T}}_{l^*i}$ but actually do not belong to the industry and $\text{FN}_i$ be the number of firms that are predicted not belonging to industry ${\mathcal{T}}_{l^*i}$ but actually belong to the industry. 
Precision $p_i$, recall $r_i$, and F1-score $F1_i$ for industry ${\mathcal{T}}_{l^*i}$ are defined as:
$$
p_i=\frac{\text{TP}_i}{\text{TP}_i+\text{FP}_i}, \  \   r_i=\frac{\text{TP}_i}{\text{TP}_i+\text{FN}_i}, \  \ 
F1_i=\frac{2 p_i r_i}{p_i + r_i}.
$$
Macro-F1 is calculated as the mean of F1-scores across industries at the focal level:
$$
\text{macro-F1}= \frac{1}{N_{l^*}} \sum_{i \in {\mathcal{T}}_{l^*}} F1_i.
$$

\subsection{Benchmark Methods} 
\label{sec:eval:bench}

We benchmark our proposed method, DeepIA, against representative existing industry assignment methods as well as prevalent methods that can be adapted for industry assignment. As discussed in Section \ref{sec:rw:autoIA}, existing methods employ machine or deep learning models for industry assignment. Therefore, we compare our method against the state-of-the-art industry assignment method proposed by \cite{wood_automated_2017}, who design a deep learning model, i.e., multi-layer perceptron, to assign firms to industries. 
We also consider other representative industry assignment methods. One method is developed based on support vector machine (SVM), a classical machine learning model \citep{roelands_classifying_2010}. Another method is developed based on ULMFiT \citep{howard_universal_2018}, which can be finetuned to do industry assignment \citep{tagarev_comparison_2019}.
While existing industry assignment methods neglect structure-based and definition-based knowledge, some recently developed classification models can be adapted to process structure-based or definition-based knowledge for industry assignment, as reviewed in Section \ref{sec:rw:class}. In this regard, we benchmark our method against a widely applied hierarchical classification method \citep{ceci_classifying_2007, silla_survey_2011}. To incorporate structure-based knowledge into classification, this method learns a SVM classifier for each node above the focal level in a tree-shaped industry hierarchy. 
Industry assignment is then conducted progressively in a top-down manner. First, the root node classifier predicts the most probable level one industry for a firm. The classier at the node corresponding to the predicted industry then selects its most probable child industry for the firm. The latter step is repeated until an industry at the focal level is assigned to the firm. In addition, we benchmark our method against 
a state-of-the-art label embedding method \citep{pappas_gile:_2019}, which can be adapted to encode definition-based knowledge. This method represents industry definitions and business description documents of firms as numerical vectors. The compatibility score between a firm and an industry is then estimated using a neural network that takes their representation vectors as inputs. Table \ref{tb:benchmethods} summarizes the methods compared in our evaluation. 

\begin{table}[h]
	\caption{Methods Compared in Our Evaluation}
	\begin{center}
		\begin{tabular}{
				L{80pt}
				L{280pt} } 
			\hline
			Method	& \multicolumn{1}{c}{Notes}  \Tstrut\\
			\hline
			DeepIA & Our proposed method  \Tstrut\\
			SVM-IA & Representative industry assignment method developed based on support vector machine (SVM) \citep{roelands_classifying_2010} \Tstrut\\
			MLP-IA & Representative industry assignment method developed based on multi-layer perceptron (MLP) \citep{wood_automated_2017} \\
			ULMFiT-IA & Representative industry assignment method developed based on ULMFiT \citep{tagarev_comparison_2019} \\
			HC-IA \Tstrut & Widely applied hierarchical classification method adapted for industry assignment \citep{silla_survey_2011} \\
			LE-IA & State-of-the-art label embedding method adapted for industry assignment \citep{pappas_gile:_2019}  \\
			\hline
		\end{tabular}
	\end{center}
	\label{tb:benchmethods}
\end{table}

Next, we discuss the implementation details of these methods. To classify a firm, each method took the firm's business description document as input and represented the document with a document embedding model (DEM). The DEM used by all the methods, except for RNN-IA, is the Doc2Vec model \citep{le_distributed_2014}. Doc2Vec was selected because it is particularly suitable for summarizing  semantics of long documents, considering that the average length of a business description document is 10,000 words.\footnote{\cite{pappas_gile:_2019} embed short documents with a sequential compositional neural network, which is not suitable to represent long business description documents in our study.}
ULMFiT-IA used its own DEM as described in \cite{tagarev_comparison_2019}. 
Definition-based knowledge used by DeepIA and LE-IA was also embedded by applying Doc2Vec to the corpus of industry definitions. 
Hyperparameters of each method were tuned and determined using validation data. 
For DeepIA, we set the embedding size $d$ to $400$. The transformation layer $e_f$ in Equation \ref{eq:firm_repr} was implemented as a multi-layer perceptron with two layers. The sizes of its input layer, hidden layer, and output layer were $400$, $800$, and $400$ respectively; the hidden layer had a dropout rate of $0.5$ and a ReLU activation function. DeepIA was trained using the Adam optimizer \citep{kingma_adam:_2015} with a learning rate of $0.001$ and a batch size of $500$.
MLP-IA had three hidden layers of sizes $640$, $4096$, and $4096$ respectively. 
For ULMFiT-IA, we follow the steps by \cite{tagarev_comparison_2019} and use $400$ as the embedding size. 
For LE-IA, we set the embedding size for both firm and industry representations to $400$. Both SVM-IA and HC-IA employed SVM with the stochastic gradient descent linear kernel. 

\subsection{Evaluation Results} \label{sec:eval:results}

Following the evaluation procedure, we set $T=2015$ and focal industry level $l^*=3$. That is, in an experimental run, each method (ours or benchmark) took firm-industry assignments and firms' business description documents from year 2012 to year 2015 as training data and learned a model to classify each new firm in year 2016 into a level three NAICS industry. Table \ref{tb:result:main} reports the average accuracy and macro-F1 for each method across $20$ experimental runs. We also list the percentage improvement by our method over a benchmark method in parentheses. 
As shown in the table, DeepIA achieves the best performance among all the compared methods in both metrics. In terms of accuracy, 68$\%$ of new firms in year 2016 are correctly classified by DeepIA, which outperforms the best performing benchmark method by 7.9$\%$. Moreover, the average macro-F1 of our method is 0.26, which is 8.3$\%$ higher than that of the best performing benchmark method. We applied  the  t-test to the performance data over 20 experimental runs and  noted  that  our  method  significantly  outperformed  each  benchmark  method  in  both metrics ($p<0.01$). We note that the values of macro-F1 are lower than those of accuracy in Table \ref{tb:result:main}. The reason is that macro-F1 is a simple average of F1-scores across industries without considering the number of firms in an industry while accuracy takes industry sizes into account. Consequently, misclassifying firms in small-sized industries has significantly negative impact on macro-F1 but relatively small impact on accuracy. We also note that macro-F1 results in our evaluation are at similar level as those reported in the literature, e.g., \cite{wood_automated_2017}.

\begin{table}[h]
	\caption{Performance Comparison between DeepIA and Benchmark Methods ($T=2015$ and $l^*=3$)}
	\begin{center}
		\begin{tabular}{l L{80pt} L{80pt}} 
			\hline
			\Tstrut & Accuracy & Macro-F1 \\
			\hline
			DeepIA  & 0.68  & 0.26  \\
			HC-IA & 0.63 (7.9\%) & 0.23 (13.0\%) \\
			LE-IA & 0.63 (7.9\%) & 0.24 (8.3\%) \\
			ULMFiT-IA & 0.62 (9.7\%) & 0.22 (18.2\%) \\
			MLP-IA & 0.61 (11.4\%) & 0.21 (23.8\%) \\
			SVM-IA & 0.60 (13.3\%) & 0.22 (18.2\%)   \\
			\hline
		\end{tabular}
	\end{center}
	\begin{center}
		\vspace{6pt}
		Note: The percentage improvement by our method over a benchmark is listed in parentheses.
	\end{center}
	\label{tb:result:main}
\end{table}

Among the compared methods, three existing industry assignment methods, SVM-IA, MLP-IA and ULMFiT-IA perform the worst because they ignore definition-based and structure-based knowledge as well as the time-specificity of assignment-based knowledge. 
In comparison to these three methods, both HC-IA and LE-IA consider additional type of knowledge (i.e., structure-based or definition-based knowledge), thereby achieving better performance. Our method attains the best performance because it integrates all three types of expert knowledge for industry assignment and takes into account the time-specificity of  assignment-based knowledge, which are realized through its two methodological innovations: dynamic industry representation and hierarchical assignment. Therefore, these methodological innovations eventually lead to the superior performance of our method.   

To ensure the robustness of our evaluation results, we performed additional evaluations by varying focal industry level $l^*$ from 2 to 5. Table \ref{tb:result:acc_otherlevels} Panel A and Panel B respectively report the average accuracy and macro-F1 for each method over 20 experimental runs. 
As reported, DeepIA remains the best method across investigated focal industry levels in both metrics. Particularly, DeepIA outperforms the best performing benchmark method by 5.6$\%$ to 9.3$\%$ in accuracy and by 8.3$\%$ to 12.0$\%$ in macro-F1.\footnote{As focal industry level $l^*$ varies from 2 to 5, the number of industries increases dramatically. Consequently, as $l^*$ increases, it becomes more difficult to accurately classify a firm to an industry and the accuracy of each method decreases (as reported in  Table \ref{tb:result:acc_otherlevels} Panel A). We do not observe exactly the same trend for macro-F1 because it is calculated differently from accuracy. Macro-F1 is a simple average of F1-scores over industries without considering the number of firms in an industry whereas accuracy takes industry sizes into consideration.} The superiority of our method over each benchmark method is also statistically significant ($p<0.01$) across investigated focal industry levels.

\begin{table}[h]
	\caption{Performance Comparison between DeepIA and Benchmark Methods ($T=2015$)}
	\begin{center}
		\begin{tabularx}{0.8\textwidth}{
				>{\raggedright\arraybackslash}p{80pt} 
				>{\raggedright\arraybackslash}X
				>{\raggedright\arraybackslash}X
				>{\raggedright\arraybackslash}X
				>{\raggedright\arraybackslash}X
				>{\raggedright\arraybackslash}X 
			} 
			\hline
			& $l^*=2$ & $l^*=3$ & $l^*=4$ & $l^*=5$ \Tstrut \\
			\hline
			\multicolumn{5}{l}{\textbf{Panel A: Accuracy}} \Tstrut
			\\
			\hline
			DeepIA & 0.70	& 0.68	& 0.57 & 0.46 \\
			HC-IA & 0.63 (11.1\%) &	0.63 (7.9\%) 	& 0.54 (5.6\%) &	0.43 (7.0\%) \\
			LE-IA & 0.64 (9.3\%)	 & 0.63 (7.9\%)  &	0.53 (7.5\%) &	0.43 (7.0\%) \\
			ULMFiT-IA & 0.63 (11.1\%)	 & 0.62 (9.7\%)  &	0.53 (7.5\%) &	0.42 (9.5\%) \\
			MLP-IA & 0.62 (12.9\%) &	0.61 (11.4\%) &	0.51 (11.8\%) &	0.42 (9.5\%) \\
			SVM-IA & 0.63 (11.1\%) &	0.60 (13.3\%) &	0.52 (9.6\%)	& 0.43 (7.0\%)  \\
			\hline
			\multicolumn{5}{l}{\textbf{Panel B: Macro-F1}} \Tstrut
			\\
			\hline
			DeepIA & 0.33 &	0.26 &	0.25 &	0.28\\
			HC-IA & 0.30 (10.0\%) &	0.23 (13.0\%) &	0.22 (13.6\%) &	0.25 (12.0\%) \\
			LE-IA & 0.28 (17.9\%) &	0.24 (8.3\%) &	0.22 (13.6\%) & 0.25 (12.0\%)\\
			ULMFiT-IA & 0.27 (22.2\%) &	0.22 (18.2\%) &	0.22 (13.6\%) & 0.25 (12.0\%)\\
			MLP-IA & 0.26 (26.9\%)	& 0.21 (23.8\%) &	0.23 (8.7\%) &	0.25 (12.0\%) \\
			SVM-IA & 0.28 (17.9\%) &	0.22 (18.2\%) &	0.20 (25.0\%) &	0.24 (16.7\%)\\
			\hline
		\end{tabularx}
	\end{center}
	\begin{center}
		\vspace{6pt}
		Note: The percentage improvement by our method over a benchmark is listed in parentheses.
	\end{center}
	\label{tb:result:acc_otherlevels}
\end{table}

We also evaluated the performance of the methods for another year.\label{pg:ay}
Specifically, we conducted an evaluation by setting $T=2013$ and focal industry level $l^*=3$. The average accuracy and macro-F1 of each method across 20 experimental runs in this evaluation are reported in Table \ref{tb:result:otheryears}. Again, our method performs the best among all the investigated methods. It outperforms the best performing benchmark method by 9.4$\%$ in accuracy and 13.6$\%$ in macro-F1 and its superiority over each benchmark method is statistically significant ($p<0.01$). 
\begin{table}[h]
	\caption{Performance Comparison between DeepIA and Benchmark Methods ($T=2013$ and $l^*=3$)}
	\begin{center}
		\begin{tabular}{l L{80pt} L{80pt}} 
			\hline
			\Tstrut & Accuracy & Macro-F1 \\
			\hline
			DeepIA  & 0.58  & 0.25  \\
			HC-IA & 0.52 (11.5\%) & 0.22 (13.6\%) \\
			LE-IA & 0.53 (9.4\%) & 0.21 (19.0\%) \\
			ULMFiT-IA & 0.53 (9.4\%) & 0.22 (13.6\%) \\
			MLP-IA & 0.51 (13.7\%) & 0.22 (13.6\%) \\
			SVM-IA & 0.52 (11.5\%) & 0.22 (13.6\%)   \\
			\hline
		\end{tabular}
	\end{center}
	\begin{center}
		\vspace{6pt}
		Note: The percentage improvement by our method over a benchmark is listed in parentheses.
	\end{center}
	\label{tb:result:otheryears}
\end{table}

To further investigate the performance of DeepIA in different contexts, we include the following analysis in appendices.
In Appendix \ref{ap:gics}, we conducted an additional evaluation with a different ICS, Global Industry Classification Standard (GICS).  
We obtained largely similar evaluation results and reported them in the appendix. 
In Appendix \ref{ap:pr}, we evaluated the performance of an industry assignment method under different degrees of automation. In other words, it is allowed to have some percentage of firms manually assigned, which emulates the real world deployment of an industry assignment method.
In Appendix \ref{ap:tree-error}, we investigated the outcome of one additional evaluation metric that accounts for partially correct predictions.
In Appendix \ref{ap:rnn-temp-agg}, we considered alternative designs of the temporal aggregation mechanism based on methods in the recurrent neural network family.
In summary, the analysis in this section and in appendices demonstrate the superiority of our method over benchmark methods in a variety of evaluation settings, which further corroborates the effectiveness of simultaneously considering the three types of expert knowledge for industry assignment.

\subsection{Performance Analysis}\label{sec:eval:ablation}

Our method features two methodological novelties: dynamic industry representation (DIR) and hierarchical assignment (HA). To evaluate the contribution of each novelty to the performance of our method, we firstly drop the HA component from DeepIA. 
As defined in Equation \ref{eq:p_abs_upwards}, the HA component computes the probability $P({\mathcal{T}}_{li}| j, t)$ of assigning firm $j$ to industry ${\mathcal{T}}_{li}$ in period $t$ recursively according to the hierarchical structure of an ICS. By dropping this component, the computation of $P({\mathcal{T}}_{li}| j, t)$ ignores the hierarchical structure of the ICS, and focuses only on industries at level $l$, i.e., industries in ${\mathcal{T}}_{l}$. Accordingly,  $P({\mathcal{T}}_{li}| j, t)$ is computed as
\begin{equation*}
	P({\mathcal{T}}_{li}| j, t)  
	=\frac{s({\mathcal{T}}_{li}, j, t)}{\sum_{i' \in {\mathcal{T}}_{l} } s({\mathcal{T}}_{li'}, j, t)}. 
\end{equation*}
We refer to the resulted method without the HA component as DeepIA-H.

Next, we further drop the DIR component. The objective of this component is to produce a dynamic industry representation $v_{li}^{(t)}$ for each industry based on its definition, structure, and assignment-based knowledge. By dropping this component, $v_{li}^{(t)}$ becomes a static industry representation, $v_{li}$, which is treated as model parameters. We refer to the resulted method with both the DIR and HA components dropped as DeepIA-HD. 

The performance difference between DeepIA and DeepIA-H reveals the contribution of the HA component to the performance of DeepIA and the performance difference between DeepIA-H and DeepIA-HD uncovers the contribution of the DIR component. To investigate the contribution of each component, we conducted experiments with 
the main experimental setting (i.e., year $T=2015$ and focal industry level $l^*=3$). Table \ref{tb:result:ablation} compares the accuracy of the three methods. 
\begin{table}[h]
	\caption{Ablation Analysis of DeepIA ($T=2015$ and $l^*=3$)}
	\begin{center}
		\begin{tabular}{l C{80pt} C{80pt} C{80pt}} 
			\hline
			\Tstrut & Accuracy & Improvement by DeepIA & Improvement by DeepIA-H \\
			\hline
			DeepIA  & 0.68 &  &\Tstrut\\
			DeepIA-H & 0.66 & 0.02 & \\
			DeepIA-HD & 0.61 & 0.07 & 0.05 \\
			\hline
		\end{tabular}
	\end{center}
	\label{tb:result:ablation}
\end{table}
As reported in the table, adding the DIR component to DeepIA-HD improves the accuracy by $0.05$ and leads to DeepIA-H, while further adding the HA component to DeepIA-H improves the accuracy by $0.02$ and leads to DeepIA. For the accuracy improvement by DeepIA over DeepIA-HD (i.e., $0.07$), $28.6\%$ of it is attributed to the HA component and the rest $71.4\%$ is contributed by the DIR component. We also conducted experiments using the metric of Macro-F1, and found that the HA component contributes $11.6\%$ of the Macro-F1 improvement by DeepIA over DeepIA-HD and the DIR component accounts for the remaining $88.4\%$. In conclusion, both the DIR and HA components contribute to the superior performance of DeepIA while DIR contributes more than HA. 

\subsection{A Case Study}
\label{sec:eval:econval}

Having demonstrated the superior performance of our method over benchmarks, it is interesting to show economic value that could be harvested from such superior performance. To that end, we conducted a case study of tax filing by firms. According to a report by Penn Wharton Budget Model (PWBM), the U.S. statutory corporate tax rate as of December 15, 2017 is $35\%$; but due to various tax deduction policies, the effective tax rate (ETR) paid by a firm is usually lower.\footnote{\url{https://budgetmodel.wharton.upenn.edu/issues/2017/12/15/effective-tax-rates-by-industry}}  Moreover, ETR varies across industries because many tax deduction policies are industry-specific. 
We obtained year 2013 ETR for each level one NAICS 2012 industry from PWBM and tabulated them in Table \ref{tb:result:etrdata}.\footnote{NAICS 2012 contains twenty level one industries. However, PWBM only provides ETR data for nineteen of them, as reported in Table \ref{tb:result:etrdata}.} For example, according to the table, year 2013 effective tax rate for a firm in the mining industry is 15.64$\%$.
\begin{table}[h]
	\caption{Year 2013 Effective Tax Rate (ETR) for each NAICS 2012 Level One Industry}
	\begin{center}
		\begin{tabular}{L{110pt} L{110pt}  L{80pt}  L{110pt}}
			\hline
			Agriculture, forestry, fishing, and hunting & Mining & Utilities & Construction \\ 
			23.82\% & 15.64\% & 24.54\% & 24.61\%   \Tstrut  \\ \hline
			Manufacturing   & Wholesale trade &  Retail trade & Transportation and warehousing  \\ 
			15.39\% & 23.73\% & 25.41\% & 26.28\%   \Tstrut  \\ \hline
			Information & Finance and insurance & Real estate, rental, leasing & Professional, scientific, technical services  \\ 
			20.57\% & 24.20\% & 24.43\% & 23.05\%  \Tstrut  \\ \hline
			Management of companies & Administrative, waste management services &  Educational services & Health care and social assistance  \\ 
			14.38\%  & 23.37\% & 26.78\% &  27.52\%  \Tstrut  \\ \hline
			Arts, entertainment, and recreation & Accommodation and food services &  Other services &  \\ 
			23.95\%  & 14.07\% & 23.73\% &    \Tstrut  \\ \hline
		\end{tabular}
	\end{center}
	\label{tb:result:etrdata}
\end{table}

We leveraged the ETR data to compute the difference between the tax amount a firm should have paid according to its true level one NAICS industry and what it would pay assuming that it filed tax according to its level one NAICS industry predicted by an industry assignment method.
Formally, for firm $j$, let $T_{j}$ be its true year 2013 ETR, $\hat{T}_{j}$ denote its year 2013 ETR according to its industry predicted by an industry assignment method, and $R_{j}$ be its taxable income in year 2013. We then calculated the misclassification cost $MC$ of a method as the average tax difference due to its classification errors:
$$
MC = \frac{1}{|{\mathcal{U}}_I|} \sum_{j \in {\mathcal{U}}_I} |T_{j}-\hat{T}_{j}| R_{j}
$$
where ${\mathcal{U}}_I$ denotes the set of investigated firms and $|{\mathcal{U}}_I|$ is the number of firms in the set. In this evaluation, we used a firm's income before extraordinary items to approximate its taxable income.

Using firm-industry assignments and firms' business description documents in year 2012 as training data, we trained each method listed in Table \ref{tb:benchmethods} to classify new firms in year 2013 to level one NAICS industries. Among $692$ new firms in year 2013, $310$ firms can be linked with valid income and ETR data and were used to compute $MC$. Firms with negative income or not belonging to any industry in Table \ref{tb:result:etrdata} (i.e., no ETR data) were dropped. The average taxable income of investigated firms is $\$573$ million. Table \ref{tb:result:econval} reports the misclassification cost $MC$ of each method in year 2013.\footnote{Because the focal industry level is one, HC-IA reduces to SVM-IA and performs the same as SVM-IA.} Among the compared methods, DeepIA incurs the lowest misclassification cost. Compared to the benchmark methods, DeepIA reduces misclassification cost by a range of $11.70\%$ to $20.09\%$. In monetary amount, it reduces tax difference by a range of $\$0.51$ million per firm to  $\$0.96$ million per firm. 

\begin{table}[h]
	\caption{Misclassification Cost of Each Method}
	\begin{center}
		\begin{tabular}{l C{80pt} C{80pt}} 
			\hline
			\Tstrut & $MC$(in millions)     & Cost Reduction by DeepIA \\
			\hline
			DeepIA  & \$3.82  &   \\
			HC-IA & \$4.71 & 18.78\% \\
			LE-IA & \$4.33 & 11.70\% \\
			ULMFiT-IA & \$4.76  & 19.75\% \\
			MLP-IA & \$4.78  & 20.09\% \\
			SVM-IA & \$4.71 & 18.78\%  \\
			\hline
		\end{tabular}
	\end{center}
	\label{tb:result:econval}
\end{table}

\section{Discussion and Conclusions} \label{sec:conclusion}

Industry assignment is fundamental to a large number of critical business practices, ranging from operations and strategic decision making by firms to economic analyses by government agencies. Existing industry assignment methods utilize only assignment-based knowledge to classify firms into industries and overlook definition-based and structure-based knowledge, although all three types of knowledge are essential for effective industry assignment. To address this limitation of existing methods, we propose a novel method that seamlessly integrates all three types of knowledge for industry assignment. Furthermore, our method considers the time specificity of assignment-based knowledge, which is also neglected by existing methods. Through extensive empirical evaluations with two widely used ICSs, we demonstrate the superiority of our method over representative existing industry assignment methods as well as prevalent classification methods that can be adapted for industry assignment. Our study contributes to the extant literature in two ways. First, our work belongs to the computational genre of design science research, which is concerned with developing computational methods to solve business and societal problems and aims to make methodological contributions \citep{rai_editors_2017, padmanabhan_machine_2022}. In this regard, the key innovations of our proposed method---dynamic industry representation and hierarchical assignment---constitute the methodological contributions of our study. Second, our study contributes to Fintech research with a novel method that effectively solves a foundational financial problem. 

\subsection{Research Implications}

This study has implications for predictive and prescriptive analytics research. Over the years, methods that predict future actions (i.e., predictive analytics) or specify optimal decisions (i.e., prescriptive analytics) have been developed to solve problems in a diverse set of domains such as health care, social networks, and Fintech \citep{abbasi_metafraud:_2012,li_utility-based_2016,fang_top_2018,fang_prescriptive_2021}. 
Our study is particular useful  for those models that rely solely on prior instance-class assignments but ignore other relevant information such as the time specificities of these assignments, the definitions of the classes, and the structural relationships among the classes. 
For instance, a problem that has a similar structure to the industry assignment problem is patent classification, which assigns a patent to a patent class \citep{gomez_survey_2014}. Three types of expert knowledge are informative for patent classification: historical patent classifications with time-stamps (i.e., dynamic assignment-based knowledge), hierarchical structure of patent classes (i.e., structure-based knowledge), and textual descriptions of patent classes (i.e., definition-based knowledge). 
Although structure-based knowledge and static assignment-based knowledge have been considered by prior patent classification methods \citep{gomez_survey_2014, aroyehun_leveraging_2021}, definition-based knowledge and time specificities of assignment-based knowledge are overlooked by these methods.
In this regard, our study informs future research about how to operationalize these three types of knowledge and incorporate them into a patent classification model.
Specifically, we can encode the three types of knowledge for patent classification with Equations \ref{eq:definition_knowledge}, \ref{eq:structure_knowledge}, \ref{eq:assign_knowledge} respectively and embed a patent filing with Equation \ref{eq:firm_repr}. Next, by applying the design of dynamic industry representation (DIR) and hierarchical assignment (HA) to patent classification, we can formulate dynamic patent class representation by embedding each patent class as a sequence of time-specific vectors that integrate the three types of knowledge, and assign a patent to a path of patent classes in accordance with the hierarchical structure entailed by the tree-shaped arrangement of patent classes.

As another example, consider the problem of link recommendation for online social networks of which the goal is to recommend friendship links to currently unlinked users \citep{li_utility-based_2016}. From the perspective of a focal user, we can construct a sequence of historical friendship links established by the user. 
This sequence bears an analogy to the dynamic assignment-based knowledge by treating the focal user as a class and the friends made by the user in the past as instances assigned to the class. 
One consequence of this perspective is that each user now has two roles: a role of class when she acts as a focal user, and a role of instance when she acts as a friend of other focal users.
Another consequence is that an instance can be assigned to multiple classes because multiple focal users can make the same friend. 
In addition to the dynamic assignment-based knowledge, definition-based knowledge about a focal user can also be built based on the user's profile. However, the definition of classes (focal users) and the description of instances (friends) now come from the same source, and can be more general than text. Lastly, a hierarchical community detection algorithm \citep{li_hierarchical_2020} can be employed to cluster focal users into hierarchical groups, which establishes the structure-based knowledge for link recommendation. In this way, while a focal user serves as a class, the communities the focal user belongs to can be regarded as meta classes.
Although the assignment-based, definition-based, and structure-based knowledge are not generated by ``experts'', our study still suggests a novel model architecture for the link recommendation problem.
Specifically, applying the design of DIR, we can construct dynamic focal user representation by embedding each focal user as a sequence of time-specific vectors that integrates the three types of knowledge about the user. Next, given that an instance can be assigned to multiple classes, the HA procedure needs to be adjusted accordingly to solve a multi-label classification problem instead of a multi-class classification problem. This can be done by normalizing each compatibility score between an instance and a class to a probability using the sigmoid function \citep{goodfellow_deep_2016}.

\subsection{Practical Implications}

Our study also has several implications for business. The analysis in Section \ref{sec:eval:econval} demonstrates the financial impact of industry assignment on firms and the government. Indeed, industry assignment can significantly affect firms' operating costs in a variety of business settings, including raising capital and purchasing commercial insurance.\footnote{Please see \url{https://www.insureon.com/small-business-insurance/cost} and \url{https://www.nav.com/blog/is-your-naics-code-costing-you-money-19128/}.} For example, an incorrect classification into a higher risk industry than the one in which a firm is actually operating can bring down the firm's business credit score, which in turn leads to more stringent loan covenants and higher interest rates. Therefore, developing methods that can produce accurate industry assignments is beneficial for both firms and the government. 
Specifically, we suggest that an automated industry assignment method like DeepIA can be deployed with a proper production rate (discussed in Appendix \ref{ap:pr}) and serve as an assistive system in the following way. Domain experts can trust its assignment predictions with top assignment scores without further intervention while focusing on verifying and correcting those predictions with low assignment scores.\footnote{A firm's industry assignment score is the probability of its most-likely assigned industry predicted by DeepIA.} By intertwining the automatic and manual industry assignment procedures, misclassification cost can be managed with minimum human effort.

In addition, our method characterizes the evolution of an industry by representing it as a sequence of time-specific vectors. Comparisons of vectors within an industry sequence or across industry sequences could shed light on how to revise the structure of an ICS in response to the changing business landscape. For instance, if the vectors for two industries gradually become similar over time, it might be appropriate to merge them into one industry. 
Finally, the three types of expert knowledge emphasized by our method are not unique to industry assignment. Patent classification is another important application where these types of expert knowledge exist and there is an urgent need to develop automated methods to classify the sheer number of patent applications into hierarchically organized patent classes \citep{gomez_survey_2014}. Therefore, our method is general and can be adapted to other critical applications where the consideration of these types of expert knowledge is essential to model performance. 

\subsection{Limitations and Future Research} \label{sec:conclusion:limit}

Our study has limitations and can be extended in multiple ways. 
First, our method represents a firm using only textual data in its business description document. Numerical data about a firm, such as financial ratios, could also convey useful information for its industry assignment because these data are correlated with the industry characteristics of the firm \citep{gupta_cluster_1972}. Future research could explore whether the incorporation of relevant numerical data into our method can further enhance its performance. 
Second, the DEM employed by our method is pretrained using the corpus of firms’ business description documents rather than learned together with other components of DeepIA. It is worth investigating whether there exists an alternative source of firms' business descriptions that convey rich information in short length, such that the DEM can be integrated into DeepIA in an end-to-end training manner.
Third, our evaluation focuses on public firms because their business description documents and industry assignment data are publicly available. It would be interesting to evaluate our method using business description documents and industry assignment data for private firms, which are usually purchased from third-party data vendors \citep{wood_automated_2017}. Considering the sheer volume of private businesses established every year, industry assignment for private firms could achieve high classification accuracy by leveraging more training data.
Fourth, our method learns industry representations using observed assignment-based knowledge but ignores future firm--industry assignment patterns. Future work could predict these patterns and represent industries with both observed assignment-based knowledge and predicted firm--industry assignment patterns. In doing so, our method could be more effective in classifying firms in a future time period.
Lastly, the learning objective of DeepIA (Equation \ref{eq:optimization}) is sequential in nature. As a result, if the method makes an incorrect decision at an upper industry level, it will misclassify a firm at lower levels. Therefore, it might be beneficial to conduct multiple label classification, which encourages the method to correctly classify a firm at all industry levels from the root to the focal one. This can be done by leveraging the factorization structure of Equation \ref{eq:p_abs_upwards}, where level-specific decision weights can be imposed to emphasize the importance of upper level decisions.

\clearpage

\bibliographystyle{informs2014} 
\bibliography{Project-autoIA} 


\renewcommand{\theequation}{\arabic{equation}} 
\ECSwitch


%

%

\begin{appendices}

\section{Additional Evaluation with GICS}\label{ap:gics}

\noindent We conducted an additional robustness check with another popular ICS -- Global Industry Classification Standard (GICS), which is jointly developed by Standard and Poor's and Morgan Stanley Capital International \citepsec{bhojraj_whats_2003_a, phillips_industry_2016_a}. 
Unlike NAICS, GICS is revised annually. However, we found that the structure of GICS was fairly stable between year $2012$ and year $2016$. Therefore, we used the version updated in year 2016 in our evaluation. Table \ref{tb:n_ind_gics} summarizes the structure of GICS 2016. As shown, the industry hierarchy of GICS 2016 contains four levels. There are 171 level four industries, each of which is denoted by a eight-digit code.
\begin{table}[h]
    \caption{Structure of GICS 2016}
    \centering
    \begin{tabular}{lcccc}
        \hline
        $l$  & 1 & 2 & 3 & 4  \\ \hline
        $N_l$ & 11 & 25 & 70 & 171 \\ 
        $D_l$ & 2 & 4 & 6 & 8  \\ \hline
    \end{tabular}
    \label{tb:n_ind_gics}
\end{table}

We collected a dataset of firm-industry assignments with GICS over the period of year 2012 to year 2016 from COMPHIST. Each record of the dataset shows the assignment of a firm to a level four GICS industry in a year. For example, \textit{Costco Wholesale Corporation} is assigned to level four GICS industry \textit{Hypermarkets and Super Centers} (coded as 30101040) in year 2012. 
Table \ref{tb:stat_GICS} reports the summary statistics of the dataset. In this table, columns $n$ and $n_{\text{new}}$ indicate the number of firm-industry assignments and the number of new firms in a year, respectively. In COMPHIST, some firms have NAICS industries but not GICS industries while some other firms carry GICS industries but not NAICS industries. Consequently, the numbers in Table \ref{tb:stat_GICS} are different from those in Table \ref{tb:stat}. 
For each firm-year observation in the firm-industry assignment dataset, we constructed the business description document of the firm as Items 1 and 1A of the firm's 10-K report filed in that year.
\begin{table}[h]
    \caption{Summary Statistics of the Firm-Industry Assignment Dataset (GICS)}
    \centering
    \begin{tabular}{lcc}
        \hline
        Year & $n$ & $n_{\text{new}}$ \\ \hline
        2012 & 6,217 & - \\ 
        2013 & 5,937 & 461 \\ 
        2014 & 5,906 & 566 \\ 
        2015 & 5,808 & 494 \\ 
        2016 & 5,455 & 310 \\ \hline
    \end{tabular}
    \label{tb:stat_GICS}
\end{table}

Following the evaluation procedure described in Section \ref{sec:eval:data}, we set $T=2015$ and focal industry level $l^*=4$. Accordingly, in an experimental run, each method listed in Table \ref{tb:benchmethods} employed firm-industry assignments with GICS from year 2012 to year 2015 as well as firms' business description documents in the same time period as training data and learned a model to classify each new firm in year 2016 into a level four GICS industry. Table \ref{tb:result:main_gics} reports the average accuracy and macro-F1 of each method across 20 experimental runs. As reported, our method achieves the best performance among all the compared methods in both metrics. It surpasses the best performing benchmark method by $7.4\%$ in accuracy and $6.1\%$ in macro-F1. Applying  the  t-test to the performance data over 20 experimental runs, we noted  that  our  method  significantly  outperformed  each  benchmark  method  in  both metrics ($p<0.01$). 

\begin{table}[h]
    \caption{Performance Comparison between DeepIA and Benchmark Methods (GICS, $T=2015$, $l^*=4$)}
    \begin{center}
        \begin{tabular}{l L{80pt} L{80pt}} 
            \hline
            \Tstrut & Accuracy & Macro-F1 \\
            \hline
            DeepIA  & 0.58  & 0.35  \\
            HC-IA & 0.53 (9.4\%) & 0.32 (9.3\%) \\
            LE-IA & 0.54 (7.4\%) & 0.33 (6.1\%) \\
            ULMFiT-IA & 0.53 (9.4\%) & 0.32 (9.3\%) \\
            MLP-IA & 0.52 (11.5\%) & 0.31 (12.9\%) \\
            SVM-IA & 0.53 (9.4\%) & 0.32 (9.3\%)   \\
            \hline
        \end{tabular}
    \end{center}
    \begin{center}
        \vspace{6pt}
        Note: The percentage improvement by our method over a benchmark is listed in parentheses.
    \end{center}
    \label{tb:result:main_gics}
\end{table}
\clearpage

\section{Performance of DeepIA under Different Production Rates} \label{ap:pr}

An automatic industry assignment (IA) method such as DeepIA can be deployed for real world applications using the concept of production rate, which refers to the percentage of firms that can be classified by the method automatically without the intervention of human experts \citepsec{gweon_three_2017_a}. 
Specifically, firms are sorted by their industry assignment scores predicted by an automatic IA method, in descending order. One example of a firm's industry assignment score is the probability of its most-likely assigned industry predicted by DeepIA. 
Given a production rate $PR\%$ of an automatic IA method, the top $PR\%$ firms are automatically assigned by the method while the rest $(1-PR)\%$ firms are manually assigned. In general, there is a trade-off between production rate and the quality of automated assignments. The higher the production rate is, the more errors tend to be in the automatically assigned group due to the inclusion of more lower quality predictions. 

Table \ref{tb:result:pr} tabulates the accuracy of our DeepIA method across different production rates ranging from $50\%$ to $70\%$, using the main experimental setting (i.e., year $T=2015$ and focal industry level $l^*=3$). As reported in the table, the accuracy of DeepIA increases from $0.73$ to $0.81$ as we reduce its production rate from $70\%$ to $50\%$. Following the practice of \cite{kearney_automated_2005}, to deploy DeepIA for a real world application, domain experts need to decide an acceptable accuracy level (e.g., $0.78$), which in turn determines the production rate of DeepIA (e.g., $60\%$). We conducted more experiments to compare the accuracy of DeepIA with that of benchmark methods across different production rates. As reported in Table \ref{tb:result:pr_bench}, DeepIA outperforms each  benchmark method significantly ($p<0.05$) across different production rates.

\begin{table}[h!]
    \caption{Accuracy of DeepIA under Different Production Rates ($T=2015$ and $l^*=3$)}
    \begin{center}
        \begin{tabular}{l L{80pt}} 
            \hline
            \Tstrut Production Rate & Accuracy \\
            \hline
            50\% & 0.81 \\
            60\% & 0.78 \\
            70\% & 0.73 \\
            \hline
        \end{tabular}
    \end{center}
    \begin{center}
        \vspace{6pt}
    \end{center}
    \label{tb:result:pr}
\end{table}

\begin{table}[h!]
    \caption{Accuracy of Each Compared Method under Different Production Rates (PR)}
    \begin{center}
        \begin{tabularx}{0.8\textwidth}{
                >{\raggedright\arraybackslash}p{80pt} 
                >{\raggedright\arraybackslash}X
                >{\raggedright\arraybackslash}X
                >{\raggedright\arraybackslash}X
            } 
            \hline
            & $PR=50\%$ & $PR=60\%$ & $PR=70\%$ \Tstrut \\
            \hline
            DeepIA &	 0.81 &  0.78 & 0.73	\\
            HC-IA & 0.78 (3.85$\%$) & 0.73 (6.85$\%$) & 0.69 (5.80$\%$) \\
            LE-IA & 0.77 (5.19$\%$) & 0.73 (6.85$\%$) & 0.68 (7.35$\%$) \\
            ULMFiT-IA & 0.76 (6.58$\%$) & 0.71 (9.86$\%$) & 0.67 (8.96$\%$) \\
            MLP-IA & 0.76 (6.58$\%$) & 0.72 (8.33$\%$) & 0.67 (8.96$\%$) \\
            SVM-IA & 0.77 (5.19$\%$) & 0.72 (8.33$\%$) & 0.67 (8.96$\%$) \\
            \hline
        \end{tabularx}
    \end{center}
    \label{tb:result:pr_bench}
    \begin{center}
        \vspace{6pt}
        Note: The percentage improvement by our method over a benchmark is listed in parentheses.
    \end{center}
\end{table}
\clearpage

\section{Tree-based Misclassification Error} \label{ap:tree-error}

By assigning a firm to an industry in a target ICS, the firm is actually assigned to the corresponding industry path leading from the root node to the assigned industry by tracing the tree-shaped hierarchy of the target ICS.  
Our evaluation in Section \ref{sec:eval:results} treats an assigned firm-industry pair as completely wrong as long as the predicted industry path does not fully overlap with the truly assigned one. However, there exists the possibility that a prediction might be partially correct by being partially aligned with the true industry path. 
To account for this possibility, we employ an additional metric of the tree-based distance \citepsec{dekel_large_2004_a, kosmopoulos_evaluation_2015_a}. 
Specifically, if the predicted industry of a firm is ${\mathcal{T}}_{l^*\hat{y}}$ while its true industry is ${\mathcal{T}}_{l^*y}$, then the tree-based distance between them is measured as the minimum number of edges from ${\mathcal{T}}_{l^*\hat{y}}$ to ${\mathcal{T}}_{l^*y}$ by viewing the given Industry Classification System (ICS) ${\mathcal{T}}$ as a tree \citepsec{kosmopoulos_evaluation_2015_a}. Consider the example ICS in Figure \ref{fig:ic_tree} of the paper. The tree-based distance between ${\mathcal{T}}_{23}$ and ${\mathcal{T}}_{24}$ is $2$ by following the path composed of edges ${\mathcal{T}}_{23} \to {\mathcal{T}}_{12}$ and ${\mathcal{T}}_{12} \to {\mathcal{T}}_{24}$, while it is $4$ between ${\mathcal{T}}_{22}$ and ${\mathcal{T}}_{24}$ by following the path composed of edges ${\mathcal{T}}_{22} \to {\mathcal{T}}_{11}$, ${\mathcal{T}}_{11} \to {\mathcal{T}}_{01}$, ${\mathcal{T}}_{01} \to {\mathcal{T}}_{12}$ and ${\mathcal{T}}_{12} \to {\mathcal{T}}_{24}$. Note that the tree-based distance is always zero from a node to itself. In this sense, a partially matched prediction has an error (measured by the tree-based distance) that is decreasing with the degree of overlap between the predicted industry path and the true industry path. Therefore, a more effective industry assignment method has a lower error.

We compared DeepIA with benchmarks in terms of the tree-based distance using the main experimental setting (i.e., year $T=2015$ and focal industry level $l^*=3$). 
For each method in an experimental run, we recorded the average of the tree-based distances across all test assignment cases. 
The evaluation results averaged across $20$
experimental runs are shown in Table \ref{tb:ap:te}. 
As reported in the table, in comparison to each benchmark method, DeepIA reduces the error measured by the tree-based distance substantially and  significantly ($p < 0.01$). 

\begin{table}[h]
    \caption{Tree-based Distance of Each Compared Method ($T=2015$ and $l^*=3$)}
    \begin{center}
        \begin{tabular}{l C{80pt} C{80pt}} 
            \hline
            \Tstrut & Tree-based Distance  & Error Reduction by DeepIA \\
            \hline
            DeepIA  & 1.58  &   \\
            HC-IA & 1.77 & 10.73\% \\
            LE-IA & 1.81 & 12.71\% \\
            ULMFiT-IA & 1.87  & 15.51\% \\
            MLP-IA & 1.93  & 18.13\% \\
            SVM-IA & 1.97 & 19.80\% \\
            \hline
        \end{tabular}
    \end{center}
    \label{tb:ap:te}
\end{table}
\clearpage

\section{Alternative Temporal Aggregation Mechanism}\label{ap:rnn-temp-agg}

In Section \ref{sec:method:dir}, we instantiate the temporal aggregation mechanism based on the multi-head self-attention mechanism (Equation \ref{eq:temp_agg_basic}). Models in the RNN family, such as LSTM, represent another paradigm of temporal aggregation: the information at each timestamp is recursively aggregated into a hidden state vector, which then becomes the context vector of the next timestamp. The recursive nature of the RNN architectures has limitations, such as precluding parallelization among training samples \citepsec{vaswani_attention_2017_a}. 
To address these limitations, the multi-head self-attention mechanism is proposed, and demonstrates superior empirical performance in tasks involving temporal aggregation \citepsec{vaswani_attention_2017_a,devlin_bert:_2019_a}. Consequently, we believe that the multi-head self-attention mechanism is the state-of-the-art architecture for temporal aggregation, and we built our DeepIA model based on it. 

To support our choice, we construct a variant of DeepIA by employing LSTM to formulate the temporal aggregation mechanism in lieu of what we report in Section \ref{sec:method:temp_agg}. 
Specifically, $v_{li}^{(t)}$, originally defined by Equation \ref{eq:temp_agg_basic}, is now a hidden state vector of a LSTM layer. The LSTM layer takes the sequence $[v_{li}^{(A,1)}, v_{li}^{(A,2)}, \dots, v_{li}^{(A,T)}]$ (computed in Section \ref{sec:method:spat_agg}) as input, and computes $v_{li}^{(t)}$ as
\begin{equation*}
    \begin{aligned}
        v_{li}^{(t)} &= \text{LSTM}(v_{li}^{(t-1)}, v_{li}^{(A,t)}) \\
    \end{aligned}
\end{equation*}
where LSTM is a standard LSTM layer provided by PyTorch. Please refer to \url{https://pytorch.org/docs/stable/generated/torch.nn.LSTM.html} for the specification of the LSTM layer. 
To incorporate the definition-based knowledge, we use $v_{li}^{(D)}$ to set the initial hidden state of the LSTM layer. 
The resulted model is named as DeepIA-LSTM. 

Next, we consider a state-of-the-art model in the RNN family: $\text{SRU}^{++}$ \citepsec{lei_when_2021_a}, which is developed based on the Simple Recurrent Unit (SRU), a strong substitute for LSTM \citepsec{lei_simple_2018_a}.
Compared to LSTM, the major design difference of SRU is a light recurrence structure and a highway network imposed on the recursively computed hidden state vectors. 
Compared to SRU, $\text{SRU}^{++}$ transforms the input sequence and then feeds the resulted sequence as the input to a SRU layer. 
We use $\text{SRU}^{++}$ to formulate the temporal aggregation in a similar way to LSTM. Specifically, we have
\begin{equation*}
    \begin{aligned}
        v_{li}^{(t)} &= \text{SRU}^{++}(v_{li}^{(t-1)}, v_{li}^{(A,t)}) \\
    \end{aligned}
\end{equation*}
where $v_{li}^{(t)}$ is now a hidden state vector of the $\text{SRU}^{++}$ layer. To incorporate the definition-based knowledge, we use $v_{li}^{(D)}$ to set the initial hidden state of the $\text{SRU}^{++}$ layer. The resulted model is named as $\text{DeepIA-SRU}^{++}$. 

We benchmark DeepIA against DeepIA-LSTM and $\text{DeepIA-SRU}^{++}$ using the main experimental setting as reported in Section \ref{sec:eval:results} (i.e., year $T=2015$ and focal industry level $l^*=3$). 
\begin{table}[h]
    \caption{Performance Comparison between DeepIA, DeepIA-LSTM and $\text{DeepIA-SRU}^{++}$ ($T=2015$ and $l^*=3$)}
    \begin{center}
        \begin{tabular}{l L{80pt} L{80pt}} 
            \hline
            \Tstrut & Accuracy & Macro-F1 \\
            \hline
            DeepIA  & 0.68  & 0.26  \\
            $\text{DeepIA-SRU}^{++}$ & 0.66 (3.0\%) & 0.25 (4.0\%) \\
            DeepIA-LSTM & 0.64 (6.3\%) &  0.24 (8.3\%) \\
            \hline
        \end{tabular}
    \end{center}
    \begin{center}
        \vspace{6pt}
        Note: The percentage improvement by our method over a benchmark is listed in parentheses.
    \end{center}
    \label{tb:ap:lstm}
\end{table}
As reported in Table \ref{tb:ap:lstm}, DeepIA outperforms $\text{DeepIA-SRU}^{++}$ by $3.0\%$ in accuracy and $4.0\%$ in macro-F1. And it surpasses DeepIA-LSTM by $6.3\%$ in accuracy and $8.3\%$ in macro-F1. All the improvements by Deep-IA are statistically significant ($p<0.01$). We note that the two variants of DeepIA inherit all the elements of DeepIA except for the implementation of the temporal aggregation component. Therefore, the performance advantages of DeepIA over $\text{DeepIA-SRU}^{++}$ and DeepIA-LSTM reflect the contribution of implementing the temporal aggregation component based on the multi-head self-attention mechanism.

\end{appendices}
%
%






\bibliographystylesec{informs2014}
\bibliographysec{Project-autoIA-ap}




%
%
%

\end{document}